\documentclass{article}

\usepackage[nonatbib,preprint]{neurips_2020}

\usepackage[utf8]{inputenc} 
\usepackage[T1]{fontenc}    
\usepackage{hyperref}       
\usepackage{url}            
\usepackage{booktabs}       
\usepackage{amsfonts}       
\usepackage{nicefrac}       
\usepackage{microtype}      

\usepackage{color}
\usepackage{bm}
\usepackage{subcaption}
\usepackage{graphicx}
\usepackage{amsmath,amssymb}

\usepackage{xspace}
\makeatletter
\DeclareRobustCommand\onedot{\futurelet\@let@token\@onedot}
\def\@onedot{\ifx\@let@token.\else.\null\fi\xspace}
\def\eg{\emph{e.g}\onedot} 
\def\ie{\emph{i.e}\onedot}

\def\etal{\emph{et al}\onedot}
\makeatother

\newcommand{\myv}{\ensuremath{\mathbf{v}}}
\newcommand{\myw}{\ensuremath{\mathbf{w}}}
\newcommand{\myW}{\ensuremath{\mathcal{W}}}
\newcommand{\myZ}{\ensuremath{\mathcal{Z}}}
\newcommand{\myWplus}{\ensuremath{\mathcal{W}^{+}}}
\newcommand{\myz}{\ensuremath{\mathbf{z}}}
\newcommand{\mySigma}{\ensuremath{\mathbf{\Sigma}}}
\newcommand{\mymu}{\ensuremath{\bm{\mu}}}
\newcommand{\myE}{\ensuremath{\mathbf{E}}}

\title{Improving Inversion and Generation Diversity in StyleGAN using a Gaussianized Latent Space}

\author{%
  Jonas Wulff \\
  MIT CSAIL \\
  \texttt{jwulff@csail.mit.edu}
  \And 
  Antonio Torralba \\
  MIT CSAIL \\
  \texttt{torralba@mit.edu}
}

\begin{document}

\maketitle
\begin{abstract}
Modern Generative Adversarial Networks are capable of creating artificial, photorealistic images from latent vectors living in a low-dimensional learned latent space. It has been shown that a wide range of images can be projected into this space, including images outside of the domain that the generator was trained on. However, while in this case the generator reproduces the pixels and textures of the images, the reconstructed latent vectors are unstable and small perturbations result in significant image distortions.
In this work, we propose to explicitly model the data distribution in latent space.
We show that, under a simple nonlinear operation, the data distribution can be modeled as Gaussian and therefore expressed using sufficient statistics.
This yields a simple Gaussian prior, which we use to regularize the projection of images into the latent space.
The resulting projections lie in smoother and better behaved regions of the latent space, as shown using interpolation performance for both real and generated images.
Furthermore, the Gaussian model of the distribution in latent space allows us to investigate the origins of artifacts in the generator output, and provides a method for reducing these artifacts while maintaining diversity of the generated images.

\end{abstract}
\section{Introduction}
Generative Adversarial Networks~\cite{Goodfellow:2014:GANs} are among the most impressive and surprising use cases of the Deep Learning revolution in computer vision in recent years.
Given only a handful of random numbers (the so-called \textit{latent vector}), these networks generate images that, to the untrained eye, are virtually indistinguishable from real photos.
This applies to indoor and outdoor settings (e.g. photos of churches or bedrooms~\cite{Karras:2019:StyleGAN}), animals such as dogs and horses~\cite{Brock:2018:Biggan}, and even to content to which human observers are highly attuned and sensitive, such as faces.
For faces in particular, StyleGAN~\cite{Karras:2019:StyleGAN} and its successor, StyleGANv2~\cite{Karras_2019_Styleganv2} have had great success in the generation of high resolution (\ie $1024 \times 1024$ pixel), high quality images, leading to entertaining past times such as games where the goal is to ``spot the fake''\footnote{For example, \texttt{www.whichfaceisreal.com} or \texttt{www.thispersondoesnotexist.com}}, but also to concerns about the authenticity and trustworthiness of photographic evidence.
In short, StyleGAN far surpasses the so-called uncanny valley and creates images that we often unquestioning accept as real.

This in turn raises the question of inversion: Given the high quality output of StyleGAN, is it possible to project an existing, real image into the latent space? That is, given an input image $I$, can we find a latent vector $\myz$ so that $G(\myz) \approx I$, where $G$ is the generator mapping the latent vector to the image?
Since directions in the latent space often correspond to semantic concepts~\cite{Karras:2019:StyleGAN,Shen:2019:InterfaceGAN,Yang:2019:SemanticHierarchies,Voynov:2020:UnsupervisedDiscovery}, a successful inversion would allow the user to perform intuitive, semantic image editing.
For example, to decrease the apparent age of a person, instead of removing every wrinkle manually, the user could simply decrease the ``age'' direction in latent space and the image would change accordingly, without the artist having to manually touch the pixels.

Due to its properties, StyleGAN is particularly well-suited for this type of application, since it first maps a random input vector $\myz \in \mathcal{Z}$ to an intermediate latent vector (or ``style'') $\myw$ in an intermediate latent space $\mathcal{W}$ using a learned mapping network $M$.
While the distribution of $\myz$ is fixed a priori, the distribution of $\myw$ in the intermediate latent space is not manually defined but is learned during training. This allows it to capture semantic directions that are specific to the data (for example, ``smile'' vs ``frown'' for human faces, which does not make sense for bedrooms), and makes the inversion process to $\myW$ easier than to $\mathcal{Z}$.
However, inverting to the latent space $\mathcal{W}$ is usually not sufficient to regenerate the input image~\cite{Abdal_2019_ICCV_Image2Stylegan,Karras_2019_Styleganv2}.
Instead, one can exploit that fact that StyleGAN uses the latent vector $\myw$ on different scales in the image generation process, and inject \textit{different} styles at different scales~\cite{Abdal_2019_ICCV_Image2Stylegan,Yang:2019:SemanticHierarchies}.
The space of such ``composite styles'', each of which consists of a separate latent vector for each scale, is usually denoted as the extended latent space $\myWplus$, and has been shown to be remarkably effective as a target space when inverting real images~\cite{Abdal_2019_ICCV_Image2Stylegan}.
Interestingly, $\myWplus$ is powerful enough to draw virtually \textit{any} image, irrespective whether it is likely under the training distribution or not.
However, those images (for example, a car drawn by a GAN trained on faces) fall into poorly behaved and unstable regions of the latent space; for example, when interpolating between such a car and a face in latent space, the generated intermediate images usually amount to nothing but noise.
When inverting to the powerful latent space $\myWplus$, inversion can therefore be seen as an ill-posed problem, and reliably solving it requires a prior on the data distribution in latent space.

In this work, we propose such a prior.
We show that, when removing the effect of the last nonlinearity in the mapping network $M$, the distribution of the resulting data can be modeled as Gaussian.
This allows us to describe the data distribution using sufficient statistics (its covariance matrix $\mySigma$ and mean $\mymu$), which in turn provide an easy-to-use prior to integrate into the inversion procedure.

Furthermore, and going beyond the problem of inversion, such a Gaussian model allows us to probe the behavior of the GAN using techniques that assume an underlying Gaussian distribution.
We demonstrate this by performing a Principal Component Analysis (PCA) on the data in the Gaussian latent space, which helps us identify when the Generator generates images that contain artifacts.
Isolating the causes of such artifacts allows us to selectively remove them without resorting to the truncation trick~\cite{Karras:2019:StyleGAN}, which, while effective, considerably decreases diversity in the output images, even for images that do not contain artifacts.

To summarize, in this work we show that (a) after a simple non-linear correction, the data distribution in the intermediate latent space of StyleGAN can be modeled as Gaussian; (b) that imposing this Gaussian as a regularization on the task of inversion leads to embeddings that are located in better parts of the latent space, as measured by interpolation between embeddings; and (c) that the Gaussian model allows us to analyze and remove artifacts of the image generator while maintaining diversity.

\subsection{Related work}
Embedding natural images into the latent space of GANs has been an active field of research. In iGAN~\cite{Zhu:2016:generative}, continuous optimization is used to obtain an embedding for an input image, serving as a starting point for interactive editing.
BiGAN~\cite{Donahue:2016:Bigan} trains an inverter simultaneously with the Generator, thereby encouraging the latent space to be well suited for inversion.
In both works, DCGAN~\cite{Radford:2015:DCGAN} is used as a Generator, and the quality is therefore limited.
Variational Autoencoders~\cite{Kingma:2013:auto,gulrajani:2016:pixelvae} suffer from a similar problem. Their focus lies on learning a latent representation of real images; while they often learn latent spaces that are semantically well behaved, their output quality is typically lower than contemporary GANs.
Furthermore, VAEs commonly impose a distribution on the latent \textit{a priori}, which can prevent them to capture modes of variation inherent in the training data, thereby reducing expressiveness. The key insight of our work is that, under an appropriate transformation, an analytical prior can be fitted to the \textit{learned} latent distribution, which gives all advantages of a prior while at the same time capturing the idiosyncrasies of the data.
Furthermore, autoencoders fix the encoder architecture during training; in contrast, our latent prior can be used with any projection method and optimization technique that can accommodate a prior.

Bau~\etal~\cite{Bau_2019_Siggraph} propose an inversion method for PGAN~\cite{Karras:2018:PGAN}, which has been shown to encode semantic concepts in the unit activations of intermediate layers~\cite{Bau:2018:Gandissect}.
They first train a separate encoder network $E(I)$, which maps an image to an approximate location in latent space.
From there, they use continuous optimization to find the best latent representation of an image; crucially, during optimization they also adjust the weights of the generator to best replicate the input image.
This procedure yields an editable latent representation of the input image and a generator that is fine-tuned to this image, which then allows modifications of real images such as the removal of windows.

Moving to higher-quality generators, the original StyleGANv2~\cite{Karras_2019_Styleganv2} proposes an inversion method to the intermediate latent space $\mathcal{W}$. Their method, using a curriculum of noise and learning rates, can invert natural images while retaining semantics (for example, the rotation of a car); however, the exact appearance often differs due to a lack of flexibility in $\mathcal{W}$.

Considering the inversion of an input image with the goal of editability, Image2StyleGAN~\cite{Abdal_2019_ICCV_Image2Stylegan} proposes to project an image into an extended latent space $\mathcal{W^{+}}$, effectively optimizing a separate style for each scale. This leads to reconstructions with high fidelity and can be used for interesting editing applications, in particular when combined with additional optimization of the noise maps~\cite{Abdal_2019_Image2StyleganPP}.
None of these works impose a data-dependent prior during the inversion process.

\section{Gaussianizing the latent space of StyleGAN}
\label{sec:space}

\newcommand{\marginalwidthleft}{0.27\textwidth}
\newcommand{\marginalwidthright}{0.205\textwidth}
\newcommand{\marginalheight}{1.85cm}
\begin{figure}[t]
	\captionsetup[subfigure]{justification=centering}
	\centering
	\begin{subfigure}[t]{\marginalwidthleft}
	\centering
		\includegraphics[height=\marginalheight]{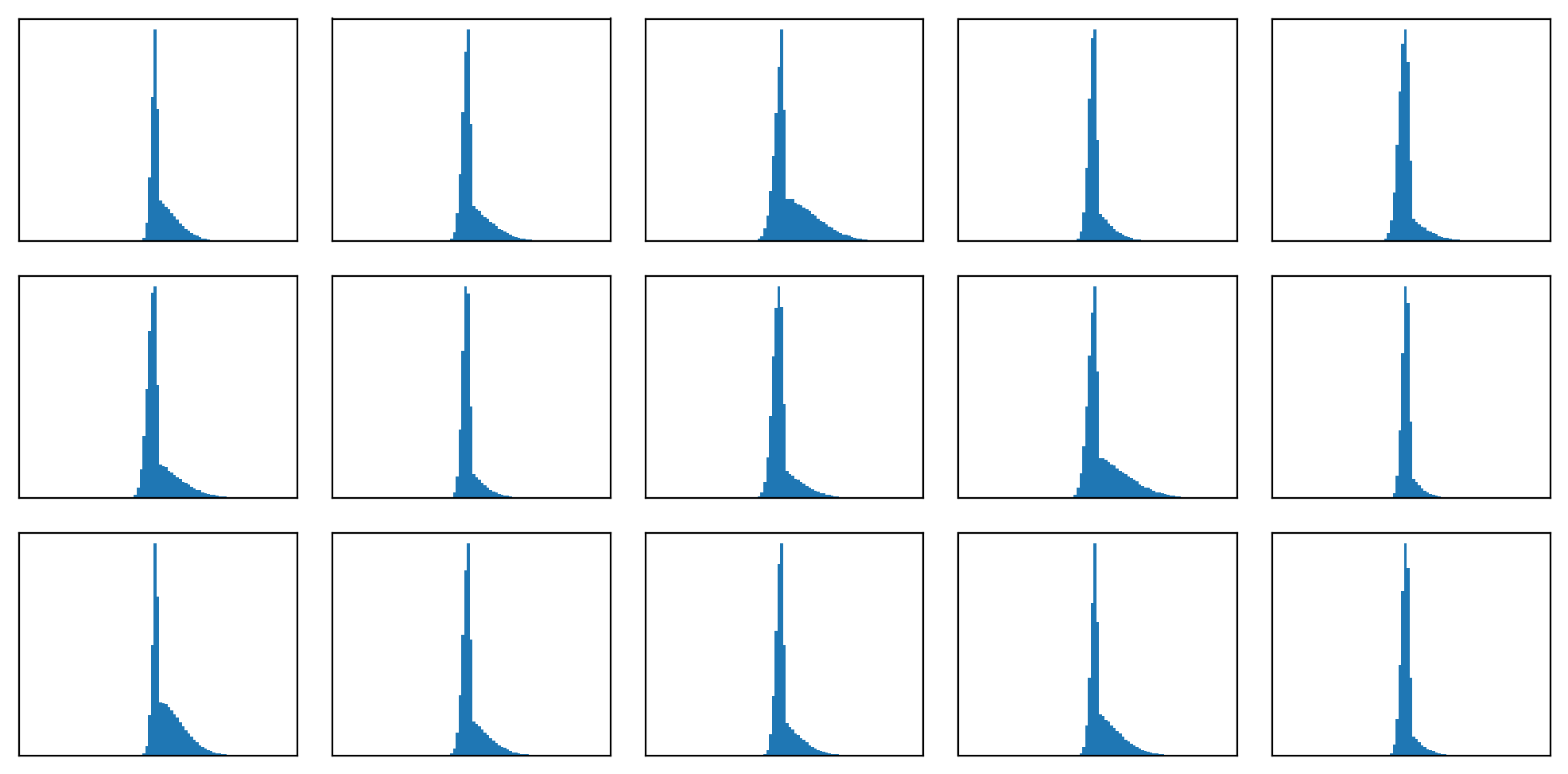}
		\caption{Marginals in $\mathcal{W}$}
	\end{subfigure}
	\begin{subfigure}[t]{\marginalwidthright}
	\centering
		\includegraphics[height=\marginalheight]{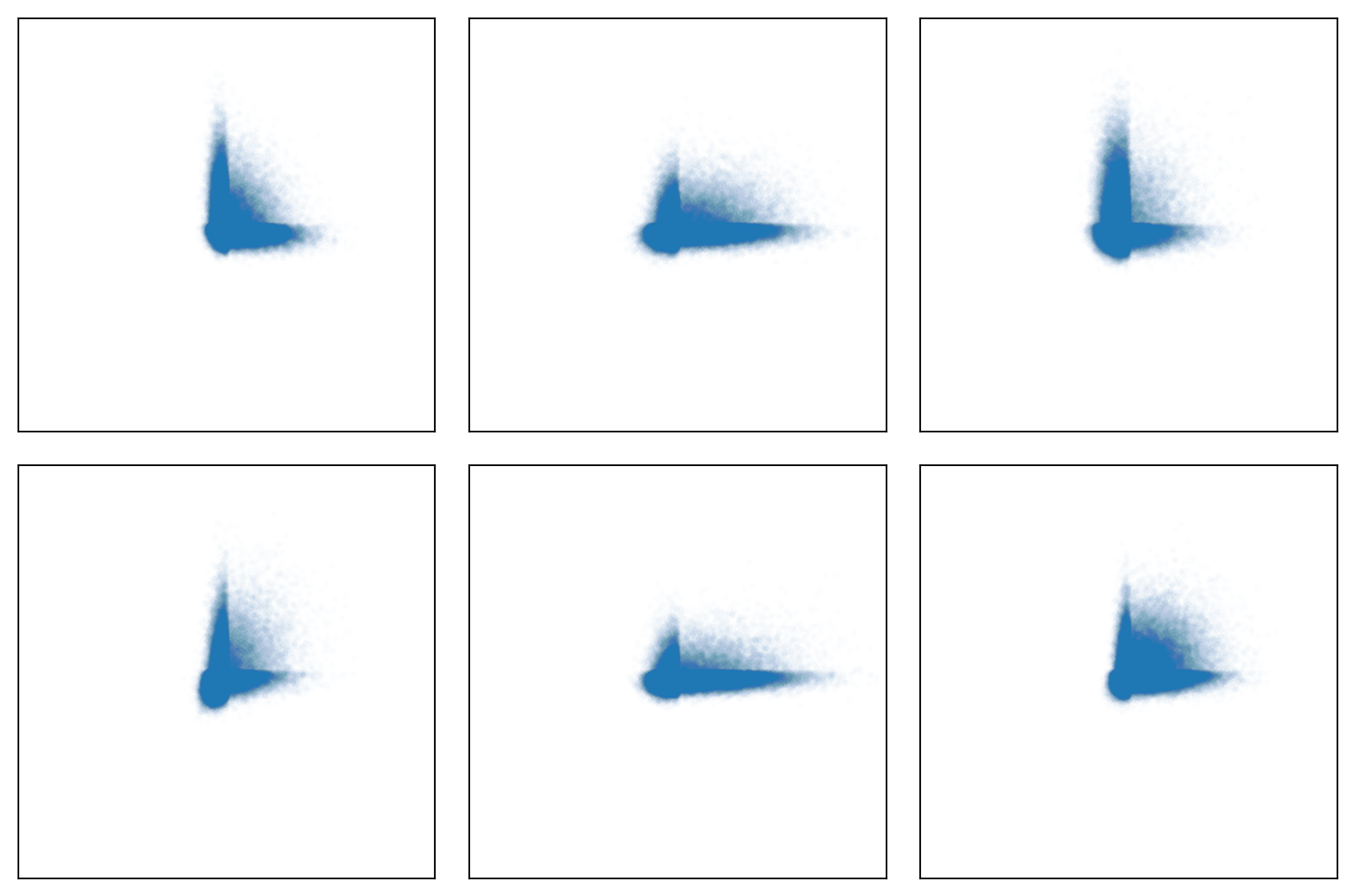}
		\caption{Pairwise distributions in $\mathcal{W}$}
	\end{subfigure}
	\begin{subfigure}[t]{\marginalwidthleft}
	\centering
		\includegraphics[height=\marginalheight]{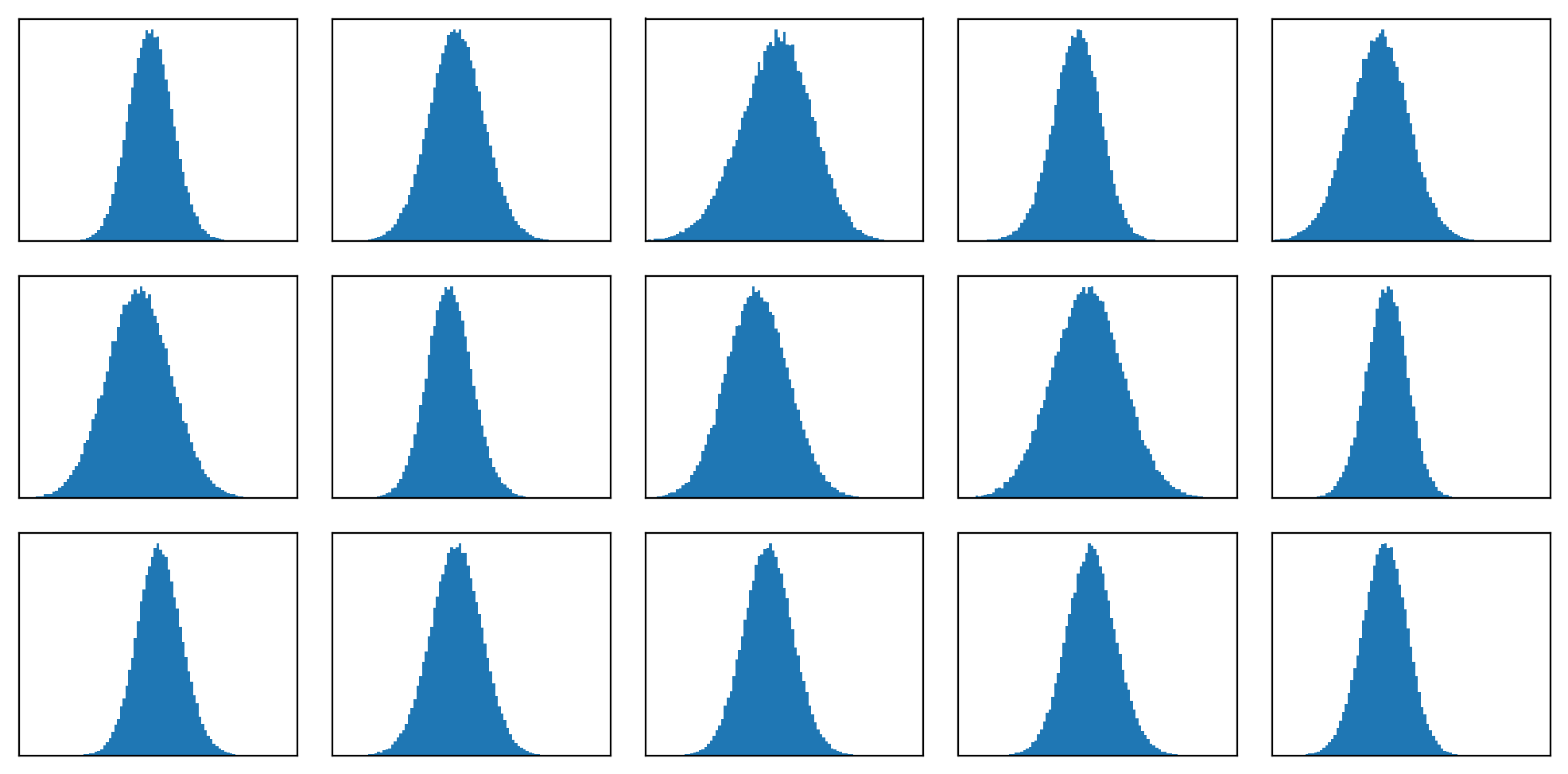}
		\caption{Marginals in $\mathcal{V}$}
	\end{subfigure}
	\begin{subfigure}[t]{\marginalwidthright}
	\centering
		\includegraphics[height=\marginalheight]{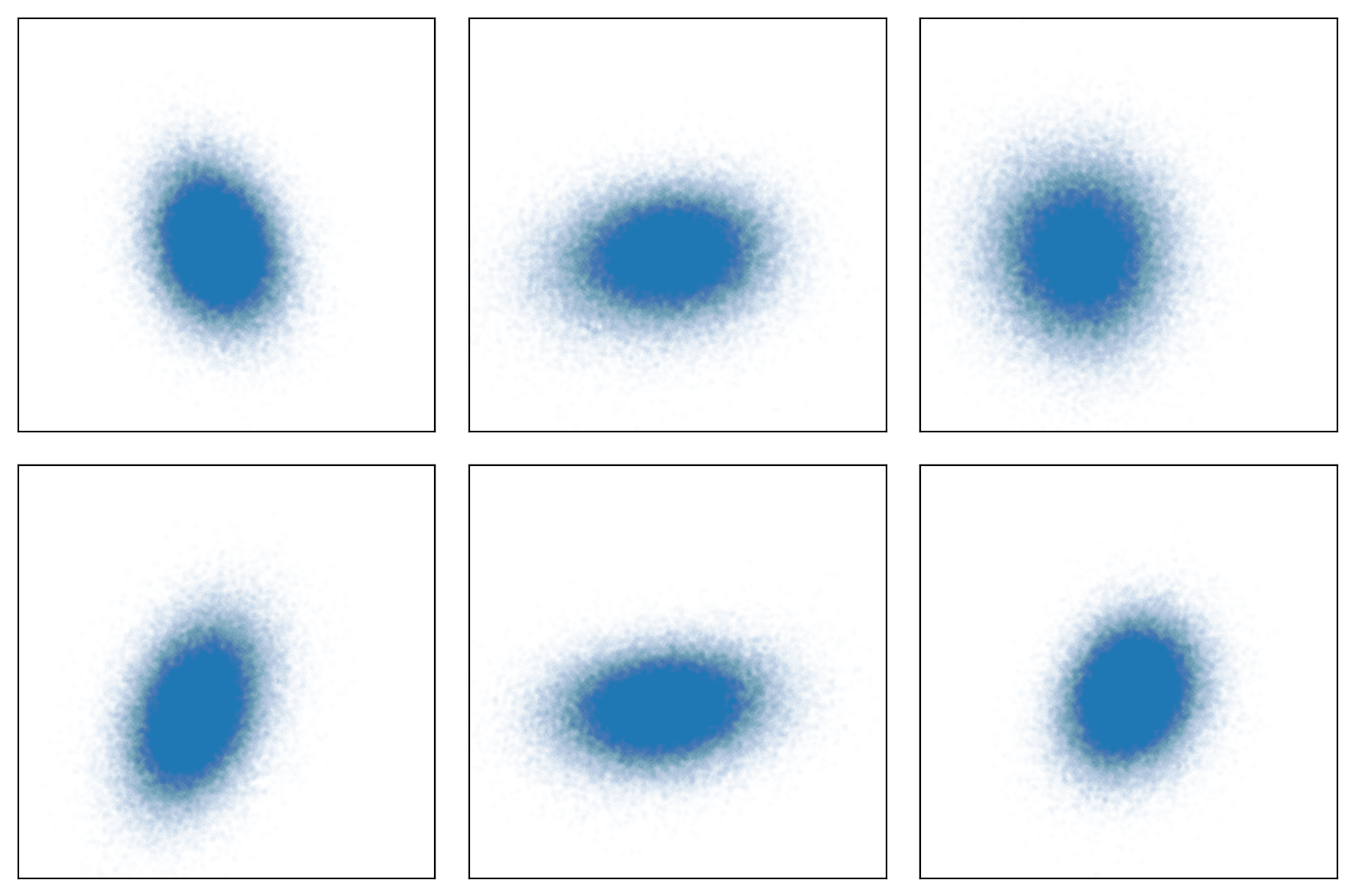}
		\caption{Pairwise distributions in $\mathcal{V}$}
	\end{subfigure}
	\caption{Statistics on $\mathcal{W}$ and $\mathcal{V}$. Marginal (a) and pairwise (b) distributions of $w$ are highly irregular. After mapping into $\mathcal{V}$, the marginal (c) and pairwise (d) distributions show that the data can be well modeled as a high dimensional Gaussian. All plots are centered at $0$ and show the same range.}
	\label{fig:marginals}
\end{figure}

To recapitulate, the processing pipeline of StyleGAN~\cite{Karras:2019:StyleGAN,Karras_2019_Styleganv2} can be roughly divided into two stages.\footnote{While StyleGAN~\cite{Karras:2019:StyleGAN} and its successor StyleGANv2~\cite{Karras_2019_Styleganv2} are separate pieces of work, their main difference lies in how they generate the output image from a given style $\myw$.
The style vector $\myw$ and how it is generated from $\myz$ does not change between both versions of StyleGAN; hence, we denote both as ``StyleGAN-based architectures''.}
First, an input latent vector $\myz \in \mathbb{R}^{d}$ is sampled from a uniform distribution on the $d$-dimensional hypersphere.
This $\myz$ is then passed through a trained mapping network $M$ which computes a style $\myw \in \mathcal{W}$.
Since no explicit constraints are imposed on the structure of $\mathcal{W}$, it learns to capture and disentangle the inherent structure of the training data~\cite{Karras:2019:StyleGAN}.
Due to this disentanglement, latent vectors $\myw$ are semantically more meaningful than $\myz$~\cite{Yang:2019:SemanticHierarchies}, and $\mathcal{W}$ is commonly used as ``the'' latent space of StyleGAN~\cite{Karras:2019:StyleGAN,Karras_2019_Styleganv2,Abdal_2019_ICCV_Image2Stylegan}.
To generate the final output image, $\myw$ is then passed to the generation network and used to modulate noise maps with increasing resolution via affine transformations of the features, resulting in the final output image.
Here, we keep the generation network fixed, and focus our attention on the structure of the latent space $\mathcal{W}$ and the distribution of $\myw$.

When inverting a real image $I$, it is typically found that inversion to $\mathcal{W}$ is easier than to $\myZ$, and common inversion methods compute $\myw \in \mathcal{W}$ so that some distance $dist(G(\myw), I)$ is minimized.
Due to the myriad possible variations of output images, however, not every image can be perfectly (or even well) recreated from a single $\myw$. Instead, the resulting regenerated image $G(\myw)$ commonly matches the semantic content of $I$, but not the exact appearance~\cite{Karras_2019_Styleganv2}.
An alternative is to use an extended latent space $\mathcal{W}^+$~\cite{Abdal_2019_Image2StyleganPP,Abdal_2019_ICCV_Image2Stylegan,Shen:2019:InterfaceGAN,Yang:2019:SemanticHierarchies}.
A point in this space is effectively a set of multiple styles, each of which is used as an input to a different scale of $G$.
Styles on different scales are thus decoupled from each other, 
greatly increasing the flexibility and coverage of the output.

However, as pointed out by Abdal \etal~\cite{Abdal_2019_ICCV_Image2Stylegan}, this increased flexibility has a curious side-effect -- the GAN can now reproduce virtually \textit{any} image, even those far outside of the domain of the training data.
For example, a GAN trained on faces has no difficulties generating a car.
However, in case of such out-of-domain inversion, the generator merely draws the pixels, but does not ``understand'' the image content, and the latent space in the immediate surrounding of the projection result is poorly behaved~\cite{Abdal_2019_ICCV_Image2Stylegan}.
For example, interpolating between such an out-of-domain image and an in-domain image rarely yields good results, instead, the intermediate images contain mostly noise.
In this work, we observe that this effect is not just relevant for out-of-domain images; instead, when inverting to $\mathcal{W}^{+}$, even images that semantically belong to the training domain (\eg faces) are often inverted to poor regions of the latent space, resulting in similar (albeit more subtle) interpolation issues.

To alleviate this issue, we propose to impose a prior on the latent vector $\myw$, thereby encouraging an inversion to stay close to the data distribution\footnote{In case of $\myWplus$, we impose our prior on each of the different styles, see the following section.}.
Unfortunately, plotting sample marginal distributions of $\myw$ (Fig.~\ref{fig:marginals}(a) and (b)) shows that the data distribution is highly irregular and hence hard to describe analytically.
Upon closer inspection of the mapping network $M$, however, we note that the very last processing step before computing $\myw$ is a Leaky ReLU (LRU) with a fixed negative slope; in the case of StyleGANv2, the slope is $\nu=0.2$.
Undoing this effect is as simple as passing $\myw$ through another LRU with negative slope $\nu = 5.0$ (from here on, we will denote Leaky ReLUs as $\text{LRU}_{\nu}$).
This yields $\myv = \text{LRU}_{5.0} \left( \myw \right)$; we denote the mapped space as $\mathcal{V}$.
Plotting marginals of the distribution of $\myv \in \mathcal{V}$ shows very strong Gaussian characteristics (Fig.~\ref{fig:marginals}(c) and (d)).
We can therefore model the distribution $p(\myv)$ on $\mathcal{V}$ as a high-dimensional Gaussian, and describe $p\left(\myv\right)$ using the empirical covariance matrix $\mySigma$ and mean $\mymu$ fitted to samples $\myv = \text{LRU}_{5.0}\left(M \left(\myz \right) \right)$.
In the remainder of this paper, we show that this Gaussian model can improve results in two independent applications, inversion of real images (Section~\ref{sec:inversion}) and removal of artifacts in generated images (Section~\ref{sec:artifacts}).

\section{Improving image inversion using the Gaussian prior}
\label{sec:inversion}
The goal of inversion is to find a point in latent space from which a given (real or generated) image can be reconstructed by the generator as accurately as possible.
If semantically meaningful directions in latent space are known, the latent can then be modified (for example to change the facial expression or camera angle) and the image re-generated with the modifications, without the user having to touch the actual pixels.
While approaches exist that directly predict the latent using a trained model~\cite{Bau_2019_Siggraph}, inversion is commonly formulated as a continuous optimization problem of the form
\begin{equation}
\hat{\myw} = \arg\min_{\myw' \in \mathcal{W}} L(I, G(\myw')),
\label{eq:inversion}
\end{equation}
where $L(\cdot)$ is some image reconstruction loss, for example the LPIPS perceptual distance~\cite{Zhang:2018:LPIPS} between the input image $I$ and the generated image $G(\myw')$~\cite{Karras_2019_Styleganv2}.

After computing the Gaussian model of $p\left(\myv\right)$ as described in the last section, it is now trivial to plug this as a prior into Eq.~\eqref{eq:inversion} by converting it to an energy term, yielding an estimate $\hat{\myw}_p$,
\begin{equation}
\hat{\myw}_p = \arg\min_{\myw' \in \mathcal{W}} L(I, G(\myw')) + \lambda \left( \myv' - \mymu \right)^{\top} \mySigma^{-1} \left( \myv' - \mymu \right), 
\end{equation}
with $\myv' = \text{LRU}_{5.0}(\myw')$ mapping from the original latent space $\mathcal{W}$ to the Gaussianized latent space $\mathcal{V}$, the empirical covariance matrix $\mySigma$ and mean $\mymu$ computed as described above, and the weight $\lambda$ determined empirically as $\lambda = 10^{-4}$.
As in~\cite{Karras_2019_Styleganv2}, in both cases (with and without the prior) we solve the inversion using ADAM as an optimizer, a learning rate of $0.1$, add ramped-down noise to the estimated latent, and run the optimization for 1000 iterations.

As a second baseline, and to be able to better invert natural images, we modify this to project into the extended latent space $\mathcal{W}^{+}$, yielding the prior-free optimization
\begin{equation}
\hat{\myw}_{(+)} = \arg\min_{\myw_{(+)}' \in \mathcal{W}^{+}} L(I, G(\myw_{(+)}')),
\end{equation}
and, adding the prior,
\begin{equation}
\hat{\myw}_{(+),p} = \arg\min_{\myw_{(+)}' \in \mathcal{W}^{+}} L(I, G(\myw_{(+)}')) + \lambda \left( \myv_{(+)}' - \mymu_{(+)} \right)^{\top} \mySigma_{(+)}^{-1} \left( \myv_{(+)}' - \mymu_{(+)} \right).
\end{equation}
Here, as above, $\myv_{(+)}' = \text{LRU}_{5.0}(\myw_{(+)}')$, $\mymu_{(+)}$ is $\mymu$ stacked $s$ times (where $s$ is the number of style scales for a given generator), $\mySigma_{(+)} = \mathbf{I}_s \otimes \mySigma$, and, in a slight abuse of notation, we allow the generator $G$ to map from both $\mathcal{W}$ and $\mathcal{W}^{+}$ to images.
The optimization parameters are the same as above, with the difference of a lower learning rate of $0.05$ and a longer optimization of 10K iterations.
Note that, since we investigate the properties of the latent space, we do not optimize the noise maps, but keep them fixed to the noise learned during training.

\subsection{Experimental setup}
We test our inversion method on four classes of images (faces, cars, horses, and churches), and use the pre-trained generator weights provided by~\cite{Karras_2019_Styleganv2}.
For each class, we use three conditions with 20 images each: real images, images generated using a single style, and images generated using different styles on different scales.
We then invert all images using the approaches described above.

\subsection{Results}
\begin{figure}[t]
	\centering
	\captionsetup[subfigure]{justification=centering}
	\begin{subfigure}[t]{0.9\textwidth}
		\includegraphics[width=\textwidth]{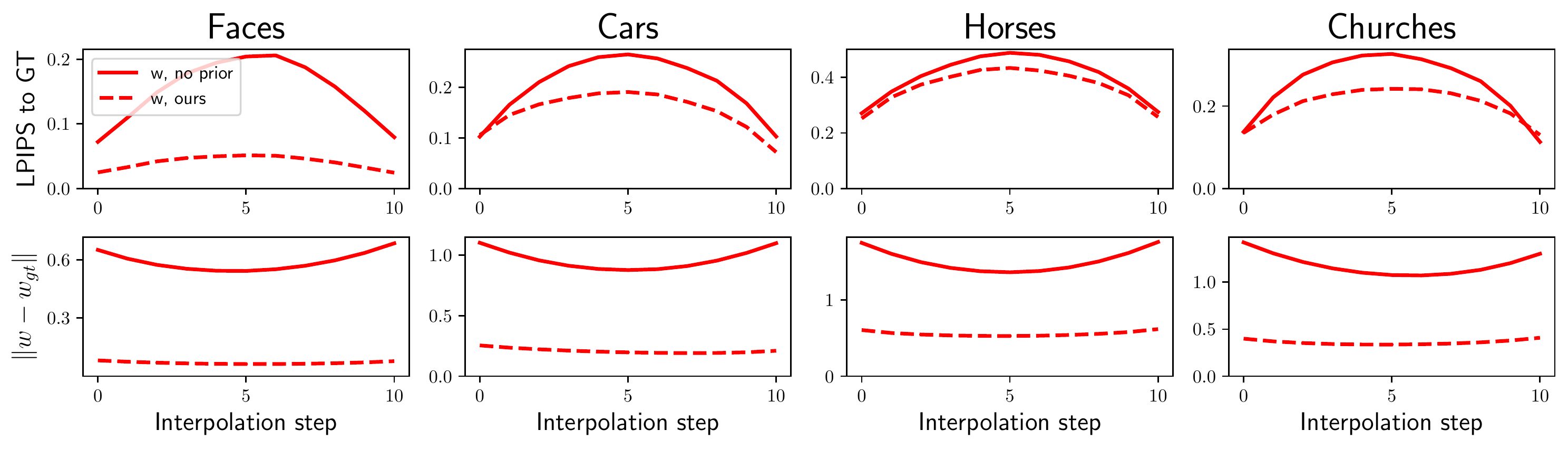}
		\caption{Reconstruction errors for images (top) and latents (bottom) when optimizing to $\mathcal{W}$.}
	\end{subfigure}
	\begin{subfigure}[t]{0.9\textwidth}
		\includegraphics[width=\textwidth]{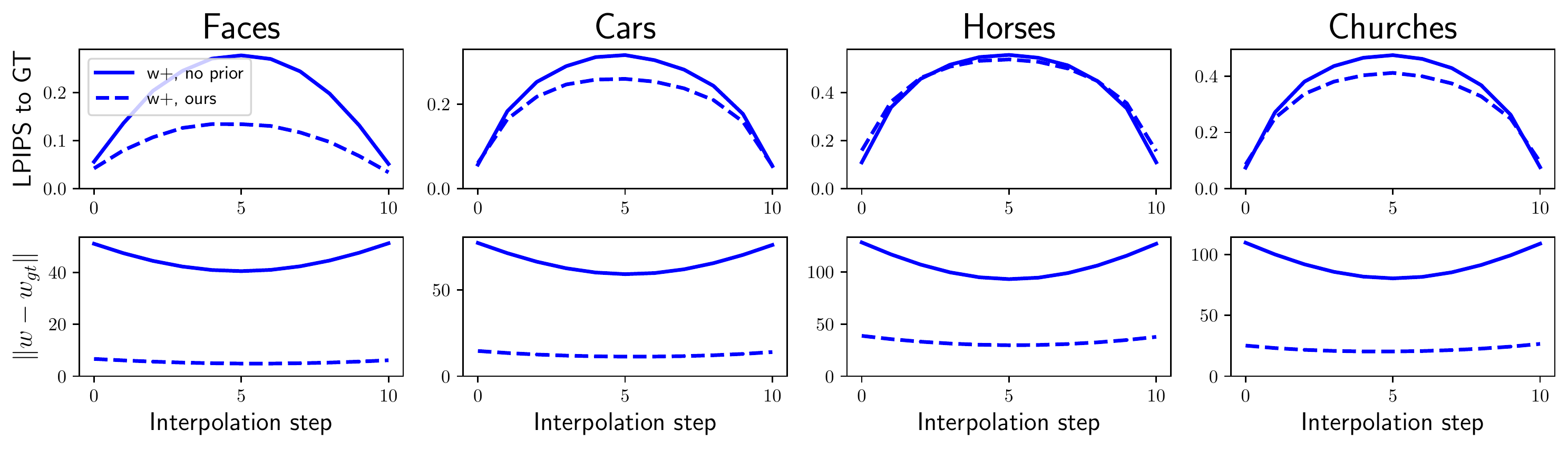}
		\caption{Reconstruction errors for images (top) and latents (bottom) when optimizing to $\mathcal{W}^{+}$.}
	\end{subfigure}
	\caption{Average image and latent reconstruction errors. Adding the prior (dashed lines) helps both when reconstructing to $\mathcal{W}$ (top, red) and $\mathcal{W^{+}}$ (bottom, blue).
		Note that we match the model to the data, \ie the images in (a) were generated using a single style, and the images in (b) were generated using different styles on different scales.}
	\label{fig:errors}
\end{figure}

\setlength{\tabcolsep}{5pt}
\begin{table}
	\caption{Reconstruction errors on real images (R), images generated from $\mathcal{W}$ (G), and from $\mathcal{W}^{+}$ (G+)
	}
	\label{tab:imageerrors}
	\centering
	\begin{tabular}{lcccccccccccc}
		\toprule
		& \multicolumn{3}{c}{Faces} & \multicolumn{3}{c}{Cars} & \multicolumn{3}{c}{Horses} & \multicolumn{3}{c}{Churches} \\
		\cmidrule(lr){2-4} \cmidrule(lr){5-7} \cmidrule(lr){8-10} \cmidrule(lr){11-13}
		\textit{Method} & R & G & G+ & R & G & G+ & R & G & G+ & R & G & G+\\
		\midrule
		$\mathcal{W}$ & 0.29 & 0.07 & & 0.22 & 0.12 & & 0.34 & 0.27 & & 0.32 & \textbf{0.11} & \\
		$\mathcal{W} + \text{prior}$ & 0.35 & \textbf{0.05} & & 0.27 & \textbf{0.04} & & 0.40 & \textbf{0.22} & & 0.39 & 0.14 & \\
		$\mathcal{W}^{+}$ & \textbf{0.15} & & 0.06 & \textbf{0.11} & & 0.06 & \textbf{0.15} & & \textbf{0.11} & \textbf{0.13} & & \textbf{0.07} \\
		$\mathcal{W}^{+} + \text{prior}$ & 0.20 & & \textbf{0.05} & 0.15 & & \textbf{0.05} & 0.22 & & 0.17 & 0.21 & & 0.09 \\
		\bottomrule
	\end{tabular}
\end{table}

\begin{table}
	\caption{Latent reconstruction errors on images generated from $\mathcal{W}$ (G) and $\mathcal{W}^{+}$ (G+)}
	\label{tab:latenterrors}
	\centering
	\begin{tabular}{lcccccccc}
		\toprule
		& \multicolumn{2}{c}{Faces} & \multicolumn{2}{c}{Cars} & \multicolumn{2}{c}{Horses} & \multicolumn{2}{c}{Churches} \\
		\cmidrule(lr){2-3} \cmidrule(lr){4-5} \cmidrule(lr){6-7} \cmidrule(lr){8-9}
		\textit{Method}		& G & G+ & G & G+ & G & G+ & G & G+\\
		\midrule
		$\mathcal{W}$ & 18.00 & & 32.53 & & 50.14 & & 37.69 & \\
		$\mathcal{W} + \text{prior}$ & \textbf{3.10} & & \textbf{5.58} & & \textbf{20.40} & & \textbf{15.66} & \\
		$\mathcal{W}^{+}$ & & 227.65 & & 316.28 & & 493.26 & & 414.96  \\
		$\mathcal{W}^{+} + \text{prior}$ & & \textbf{31.02} & & \textbf{59.64} & & \textbf{151.23} & & \textbf{101.55} \\
		\bottomrule
	\end{tabular}
\end{table}

\textbf{Quantitative results.}
Table~\ref{tab:imageerrors} shows the image reconstruction errors~\cite{Zhang:2018:LPIPS} for real and generated images; for generated images, Table~\ref{tab:latenterrors} shows the $L_{2}$ distances in latent space between the estimated latent and the true latent.
Note that, when using generated images, we match the model to the data, \ie when inverting images that were generated from a single style, we invert to $\mathcal{W}$; otherwise, we invert to $\myWplus$.
As can be seen in Table~\ref{tab:imageerrors}, our method slightly decreases raw reconstruction performance for real images; this is to be expected, since without a prior, the model overfits to the data and hence achieves better reconstruction.
However, when comparing the error of the latent
in Table~\ref{tab:latenterrors}, our method finds latents that are significantly closer to the true latents.

Figure~\ref{fig:errors} shows the implications of this. Here, we sample 80 pairs per condition and class, interpolate between the generated images using the ground truth latents, and compare these ground truth interpolations to the interpolations between the estimated latents.
As can be seen, imposing the prior significantly improves performance in the middle points of the interpolations, both in case of optimizing to $\mathcal{W}$ (red) and optimizing to $\mathcal{W}^{+}$ (blue). That this is the case even for the classes ``Horse'' and ``Church'' in Fig.~\ref{fig:errors}(b), where the prior causes worse inversion performance for the start and end images, indicates that the prior causes the inversion to find points in the latent space that are better behaved and overfit less to the input data.

\textbf{Qualitative results.}
\newcommand{\examplewidth}{1.0\textwidth}
\begin{figure}[t]
	\centering
		\includegraphics[width=\examplewidth]{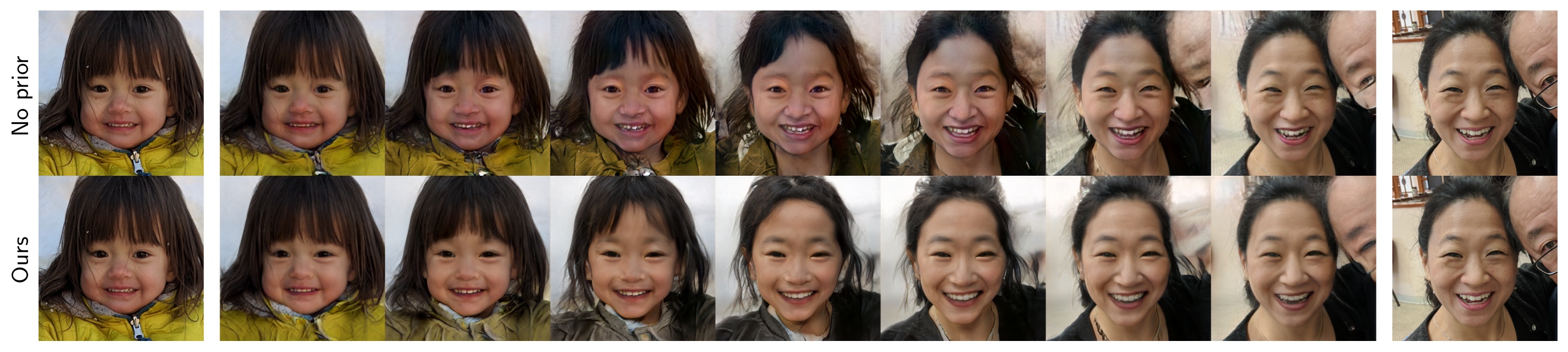}
		\includegraphics[width=\examplewidth]{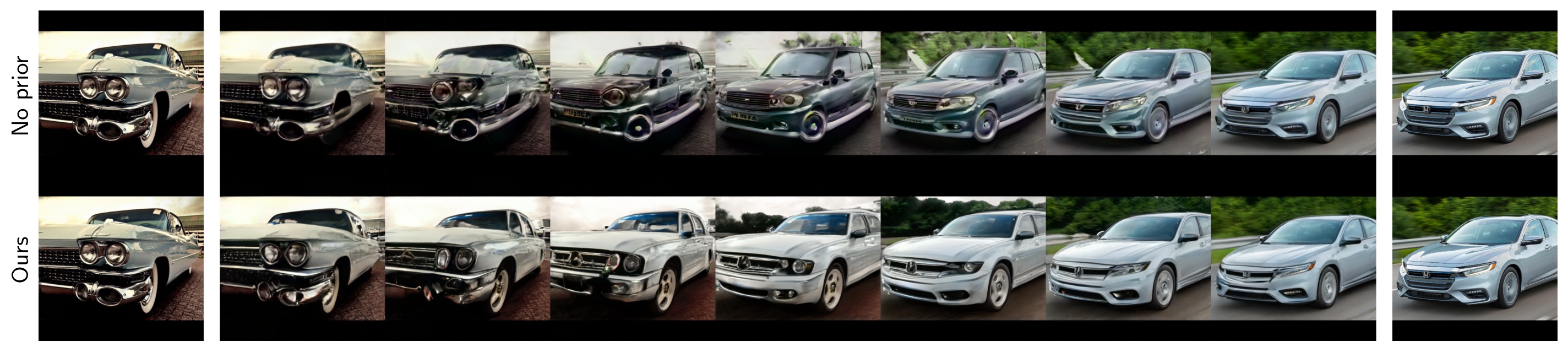}
		\includegraphics[width=\examplewidth]{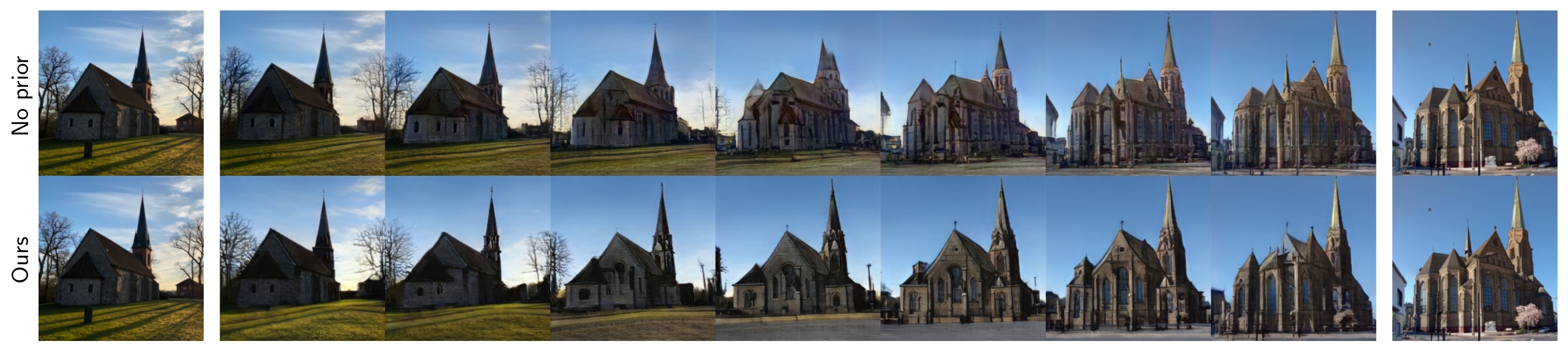}
		\includegraphics[width=\examplewidth]{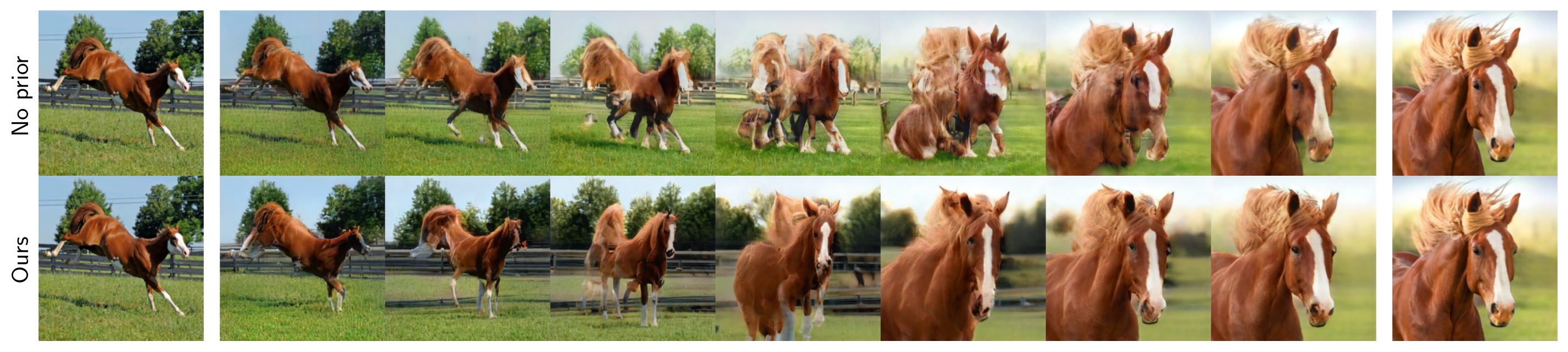}		
		\caption{Example interpolations between inversions of real images to $\mathcal{W}^{+}$. Without a prior (top row in each example), the latents often fall into poorer areas of the latent space, causing distorted appearances in the intermediate images. Using a prior (bottom rows) encourages the latent to lie in good regions, causing the intermediate images to be more realistic. The left and right columns show the start and end frames, respectively. Please see the appendix for additional examples.}
		\label{fig:interpolationexamples}
		\end{figure}
Figure~\ref{fig:interpolationexamples} shows this effect visually on interpolations between real images. As can be seen, inverting without our prior often leads to interpolations that contain unrealistic appearances, shapes, and color artifacts (top rows); when inverting with the prior (bottom rows), the interpolations stay realistic.

\begin{table}
	\caption{User study of interpolation quality (given in percent preference)} 
	\label{tab:userstudy} 
	\centering 
	\begin{tabular}{lcc} 
		\toprule 
		\textit{Dataset} & Inversion without prior & Inversion with prior (ours) \\
		\midrule
		Faces & 10.1\% & \textbf{89.9\%} \\ 
		Cars & 34.2\% & \textbf{65.8\%} \\ 
		Horses & 32.6\% & \textbf{67.4\%} \\ 
		Churches & 42.9\% & \textbf{57.1\%} \\ 
		\midrule
		All & 29.8\% & \textbf{70.2\%} \\
		\bottomrule 
	\end{tabular} 
\end{table}     

\textbf{User study.}
To quantify the impact of the prior on interpolation quality, we perform a user study.
We first invert all real images (20 each for faces, cars, horses, and churches) to $\myWplus$, both with and without prior.
For all pairs within the same condition (with or without prior), we then compute the interpolated center image by generating the image from the midpoint of their $\myw_{(+)}$.
The center images for inversions with and without prior are then shown side-by-side in shuffled order, and users are asked which of the two images they judge to be more realistic.
In total, $n=22$ subjects took part in the study, resulting in 1388 judgements.
Table~\ref{tab:userstudy} shows the results. In all cases, interpolations from inversions with prior (\ie our method) are preferred over interpolations from inversions without prior, and overall, our prior is preferred in 70\% of all cases.
For faces, around 90\% of users prefer our prior.
For the other datasets, preference is somewhat lower. We believe this is because the quality of generated images is lower for cars, horses, and churches, occasionally resulting in unrealistic interpolations for both methods. Furthermore, in some instances it is hard to judge the realism, for example, whether a church has zero, one, or two spires does not influence how realistic the image is perceived to be.

\section{Removing image artifacts using the Gaussian prior}
\label{sec:artifacts}

\newcommand{\samplewidth}{0.32\textwidth}
\begin{figure}[t]
	\centering
	\begin{subfigure}[b]{\samplewidth}
		\includegraphics[width=\textwidth]{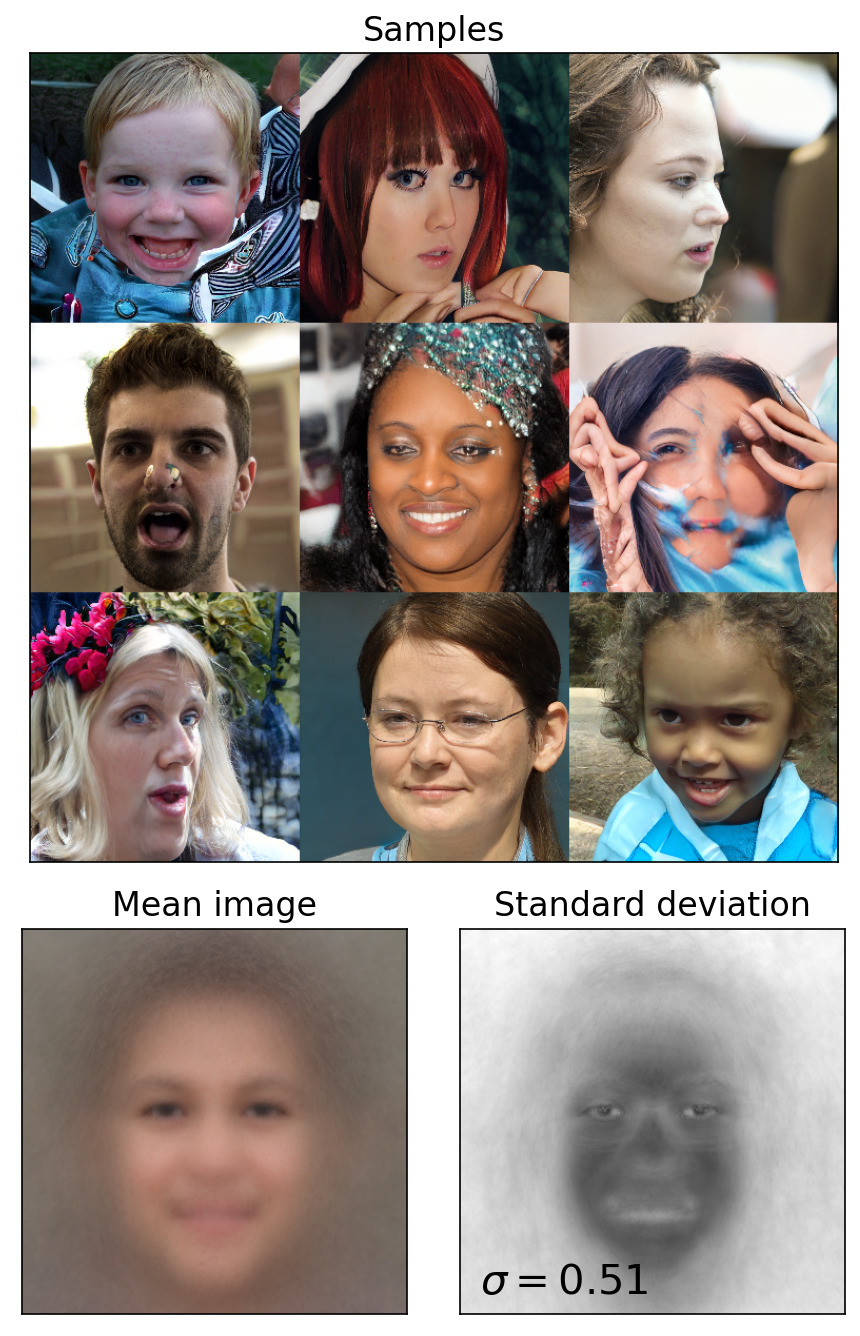}
		\caption{Uncorrected output}
	\end{subfigure}
	\hspace{0.02in}
	\begin{subfigure}[b]{\samplewidth}
		\includegraphics[width=\textwidth]{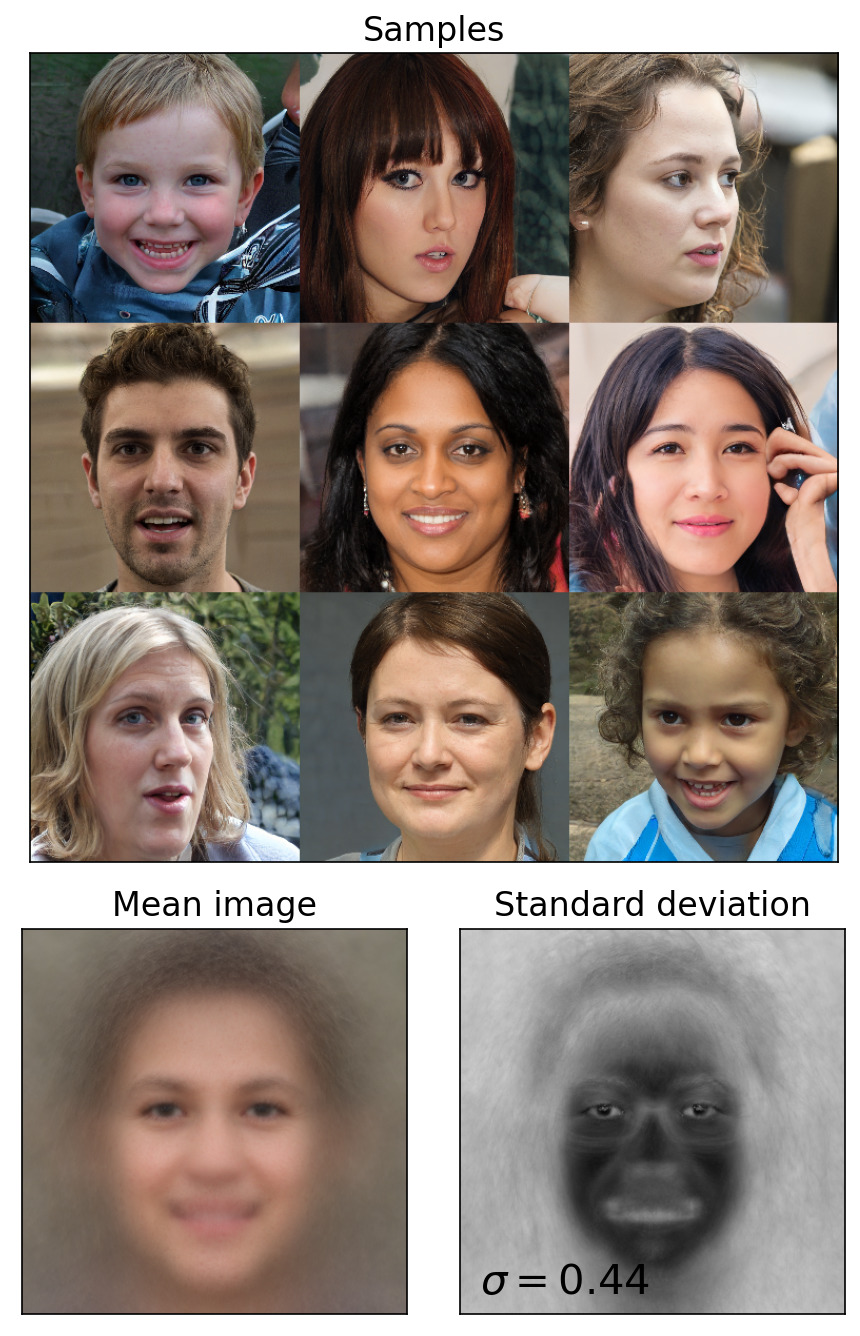}
		\caption{Truncation, $\psi=0.7$}
	\end{subfigure}
	\hspace{0.02in}
	\begin{subfigure}[b]{\samplewidth}
		\includegraphics[width=\textwidth]{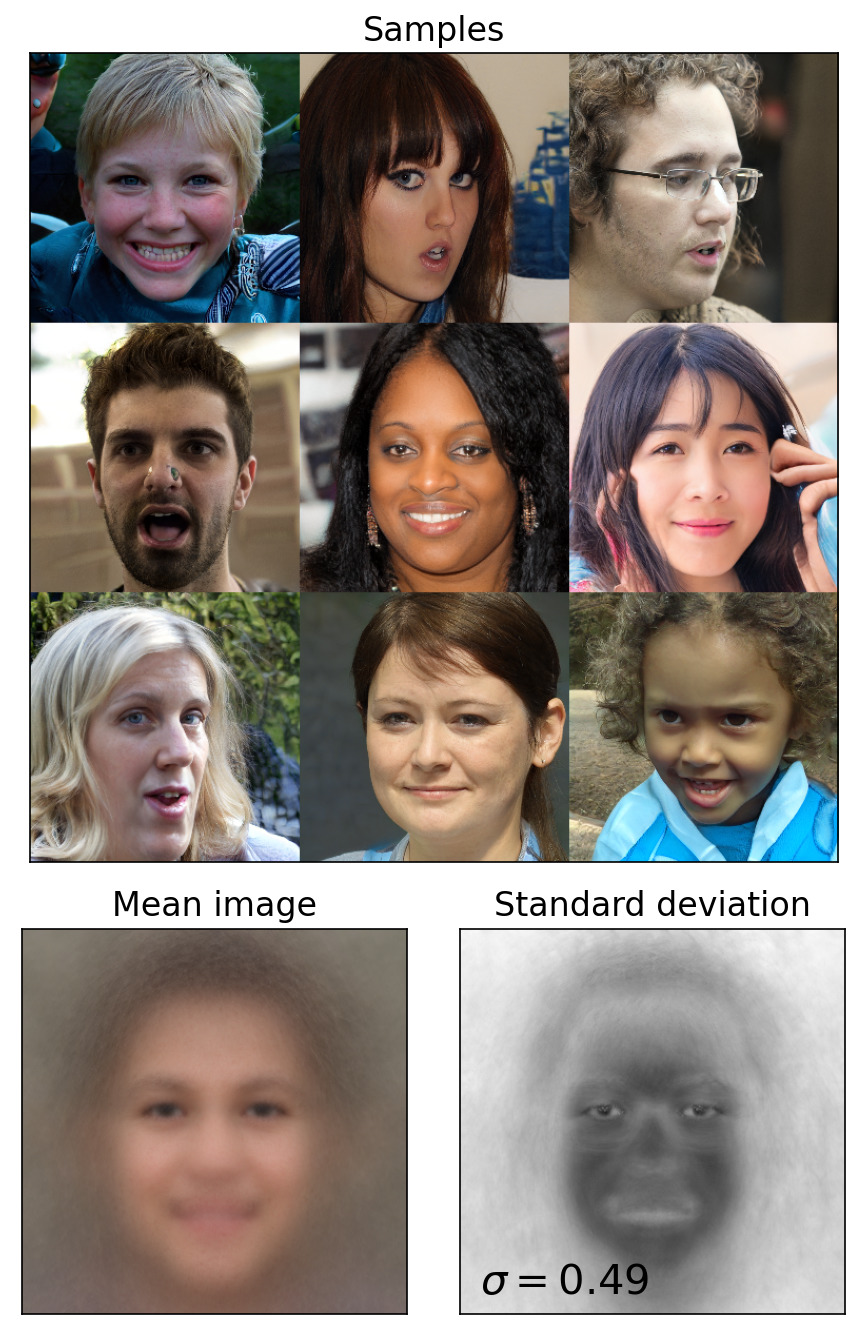}
		\caption{Ours, $\tau = 0.5$}
	\end{subfigure}
	\caption{Comparison of different correction methods. Raw samples (a) are commonly corrected using truncation (b), which removes artifacts, but also reduces diversity, as can be seen from the sharper average image and lower per-pixel standard deviations in (b). Our method (c) reduces artifacts, while maintaining a high degree of diversity.}
	\label{fig:samples}
\end{figure}

Leaving the subject of inversion behind, we ask whether our Gaussian model can help us better understand the behavior of the GAN.
Modeling the data distribution in the latent space as Gaussian allows us to probe this behavior using methods that assume that the underlying data is Gaussian distributed, such as Principal Component Analysis (PCA).
Here, we show how this helps us to identify and remove artifacts from generated images while maintaining diversity, which is crucial in the context of face image generation.

\subsection{Analyzing artifacts}
While StyleGAN-based architectures generally produce outputs of a high visual quality, they often still contain a number of artifacts.
In Fig.~\ref{fig:samples}(a), examples are the distorted occluder of the face (middle row, right) or artificial flower-like patches on the head (bottom row, left).

A common trick to improve visual appearance and remove such artifacts is \textit{truncation}\footnote{Note that, mathematically, this process does not truncate individual values, but compresses them towards the mean. However, for consistency with previous work, we keep the terminology introduced in~\cite{Karras:2019:StyleGAN}}.
For a given input $\myz$ and its corresponding $\myw = M\left(\myz\right)$, a truncated latent $\tilde{\myw}$ is computed by moving $\myw$ towards the empirical mean $\bar{\myw}$ by a factor of $\psi$,
\begin{align}
\tilde{\myw} = \bar{\myw} + \psi \left( \myw - \bar{\myw} \right) = \psi \myw + \left( 1 - \psi \right) \bar{\myw}.
\end{align}
The resulting $\tilde{\myw}$ is then used as the input to the generator network.
As shown in Fig.~\ref{fig:samples}(b), this removes virtually all of the artifacts.
At the same time, visual diversity is reduced.
The generated faces are closer to a canonical, frontal pose, the lighting is very flat, there is much less facial hair, and the skin is lighter.
This can also be seen when comparing the mean images and per-pixel standard deviations (Fig.~\ref{fig:samples}, bottom); when using truncation, the mean image is sharper and the per-pixel standard deviation is lower, especially in regions where skin is visible.
This is obviously a problem when trying to capture the diversity of humans and their appearances.

\newcommand{\pceffectswidth}{0.48\textwidth}
\begin{figure}[t]
	\captionsetup[subfigure]{justification=centering}
	\centering
	\begin{subfigure}[t]{\pceffectswidth}
		\centering
		\includegraphics[width=\textwidth]{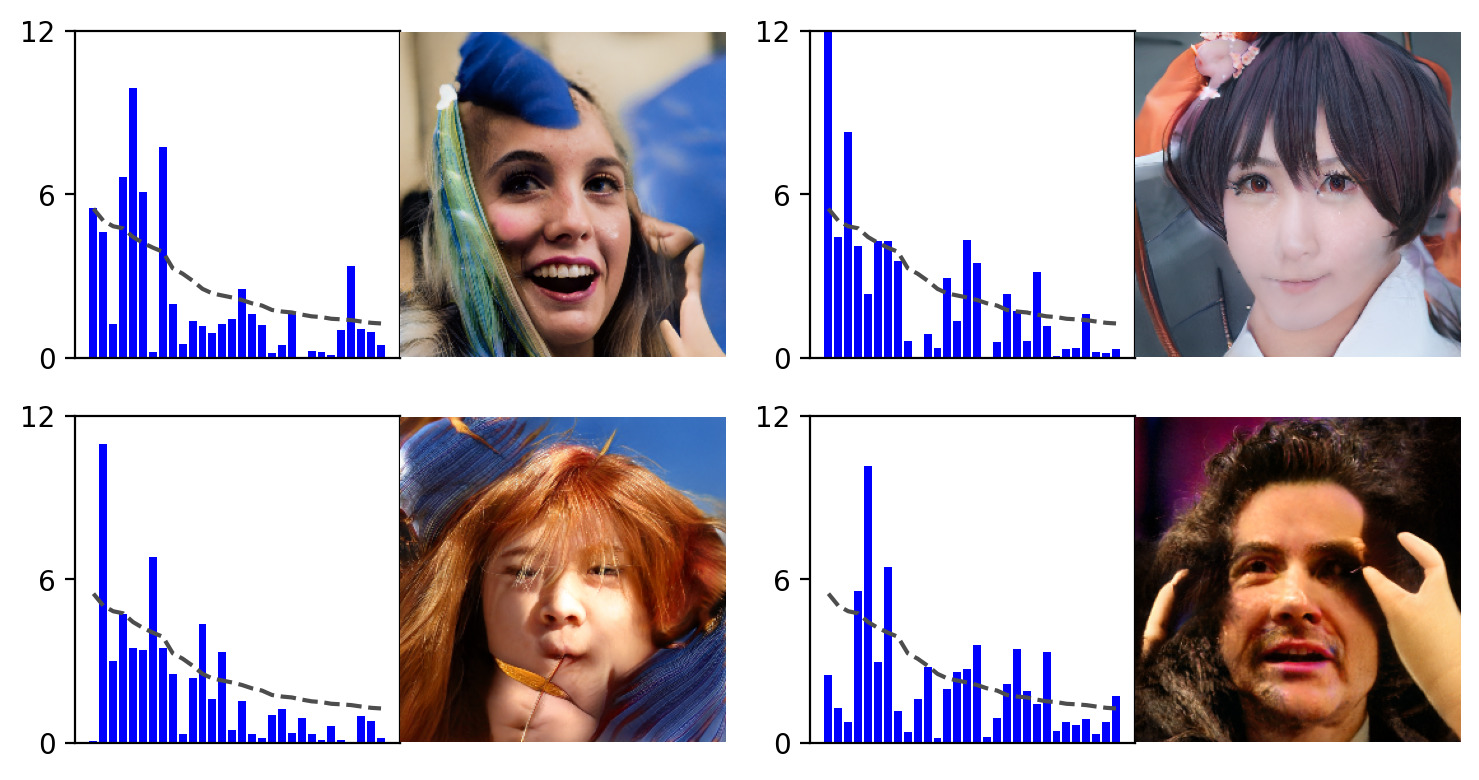}
		\caption{Images with artifacts and magnitudes of their first 30 principal components}
		\label{fig:pc_comparison_bad}
	\end{subfigure}
\hspace{0.1in}
	\begin{subfigure}[t]{\pceffectswidth}
	\centering
	\includegraphics[width=\textwidth]{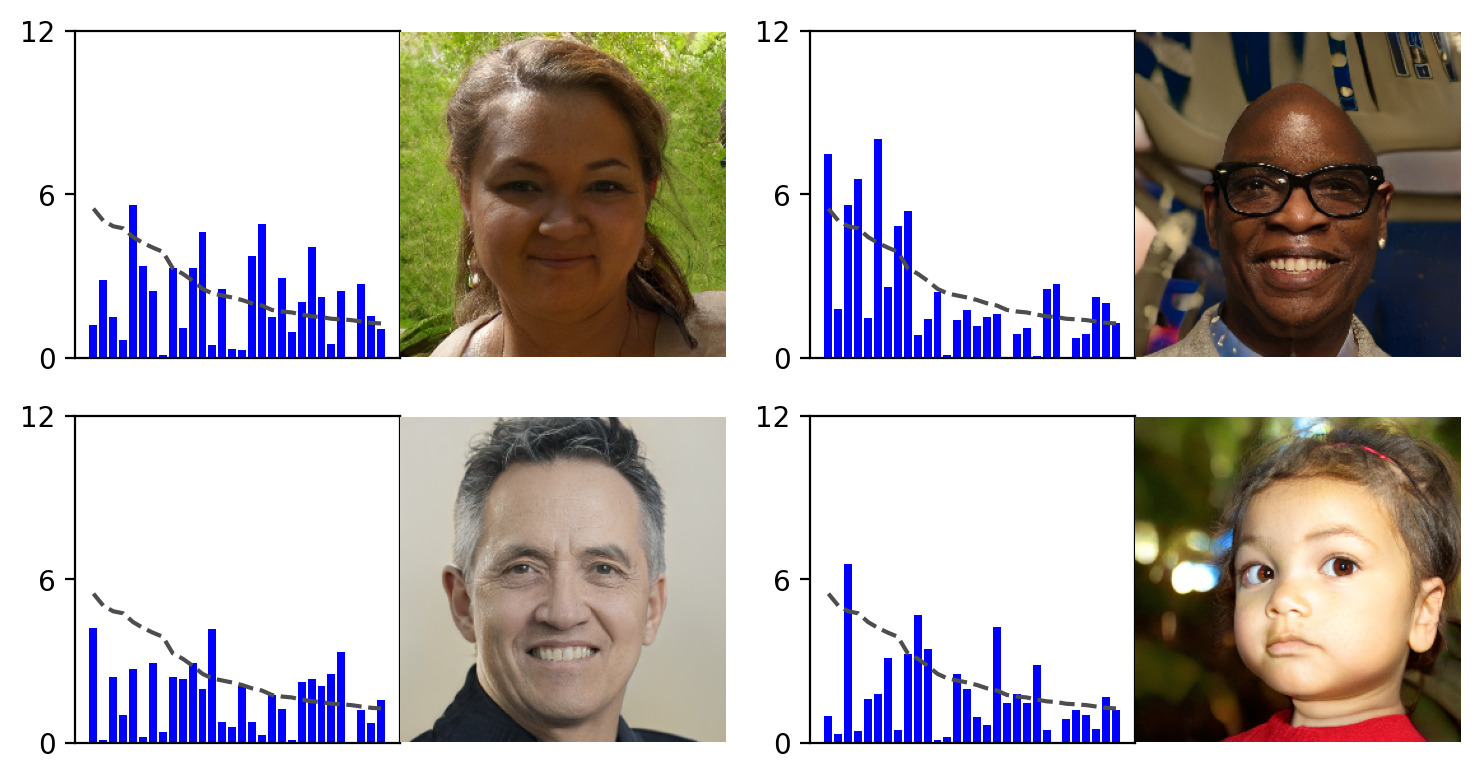}
	\caption{Images without artifacts and magnitudes of their first 30 principal components}
	\label{fig:pc_comparison_good}
\end{subfigure}
	\caption{Principal components of images with and without artifacts. Images with artifacts exhibit significantly larger magnitudes in the low principal components than good images; the dashed lines indicate one standard deviation.}
	\label{fig:pc_comparison}
\end{figure}

To break this tradeoff between artifacts and diversity, we investigate if it is possible to remove artifacts in a targeted manner, leaving images without artifacts untouched.
Since we model $p\left(\myv\right)$ as Gaussian, we can compute the main axes of variation in $\mathcal{V}$ using Principal Component Analysis (PCA), which assumes the underlying data to be Gaussian distributed.
Projecting latents into the PCA space (we denote the projection of $\myv$ as $\myv^{p}$) reveals that latents that generate images with artifacts exhibit significantly larger values, especially in the lowest dimensions (\ie those with the largest variation).
Figure~\ref{fig:pc_comparison} shows a comparison of the values in the first $30$ dimensions of $\myv^{p}$ for images with artifacts (Fig.~\ref{fig:pc_comparison_bad}) and without artifacts (Fig.~\ref{fig:pc_comparison_good}).
Note that the latents generating images with artifacts are still valid samples from a Gaussian; the artifacts are merely hiding in the tails of the distribution.

\subsection{Reducing artifacts by logarithmic compression}
To reduce the impact of these large components, we propose to logarithmically compress values with a magnitude larger than a threshold $\tau \sigma$, where $\tau$ is a scaling factor and $\sigma = \max_i \sigma_i = \max_i \sqrt{\lambda_i}$ is the maximum standard deviation along the principal components, corresponding to the square root of the maximum eigenvalue of $\mySigma$,
\begin{equation}
\tilde{v}^p_i =
\begin{cases}
	\text{sign} \left( v^p_i \right)
		 \tau \sigma \left[ \log \left( \frac{ \left| v^p_i \right| }{\tau \sigma} \right) + 1 \right] & \text{ if } ~ \left| v^p_i \right| > \tau \sigma \\
v^p_i & \text{ otherwise },
\end{cases} 
\end{equation}
where $v^p_i, \tilde{v}^p_i$ are the $i$-th component of $\myv^p$ and $\tilde{\myv}^p$, respectively.
After thus computing $\tilde{\myv}^p$, we reproject $\tilde{\myv}^p$ into $\mathcal{W}$ as $\tilde{\myw} = \text{LRU}_{0.2}\left( \myE \tilde{\myv}^p + \mymu \right)$, where the columns of $\myE$ contain the eigenvectors of $\mySigma$, and generate the image from $\tilde{\myw}$ as before.

\newcommand{\samplecomparisonwidthgraph}{0.39\textwidth}
\newcommand{\samplecomparisonwidthoriginal}{0.1\textwidth}
\newcommand{\samplecomparisonwidthfid}{0.2\textwidth}
\newcommand{\samplecomparisonheight}{6cm}
\newcommand{\samplecomparisongraphheightparam}{2.4cm}
\newcommand{\samplecomparisongraphheightface}{3.6cm}
\begin{figure}[t]
	\captionsetup[subfigure]{justification=centering}
	\centering
	\begin{subfigure}{\samplecomparisonwidthgraph}
		\centering
		\begin{subfigure}[t]{\textwidth}
			\centering
			\includegraphics[height=\samplecomparisongraphheightparam]{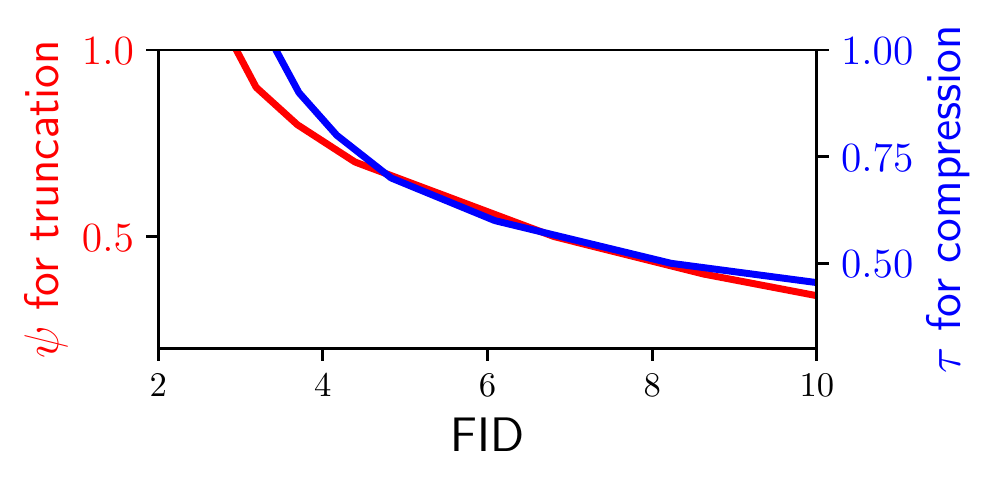}
			\vspace{-0.1in}
			\caption{Correction parameters vs. FID}
		\end{subfigure} \\ \vspace{0.1in}
		\begin{subfigure}[t]{\textwidth}
						\centering
			\includegraphics[height=\samplecomparisongraphheightface]{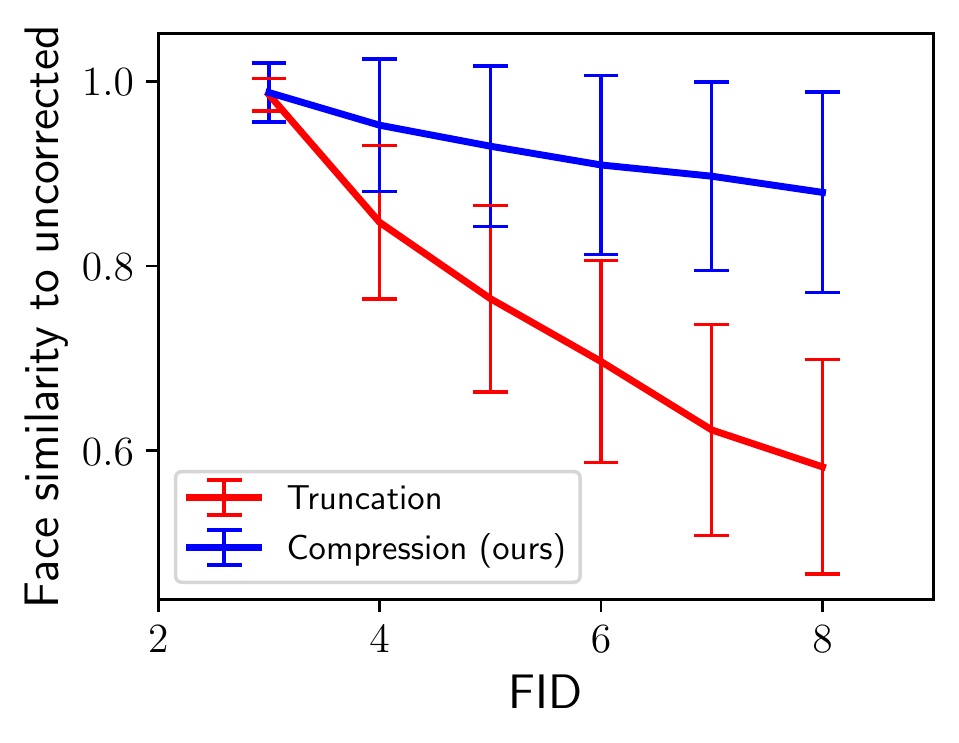}
			\vspace{-0.1in}
			\caption{Face distance vs. correction strength}
		\end{subfigure}	
	\end{subfigure}
	\hspace{0.1in}
	\begin{subfigure}{\samplecomparisonwidthoriginal}
		\centering
		\includegraphics[height=\samplecomparisonheight]{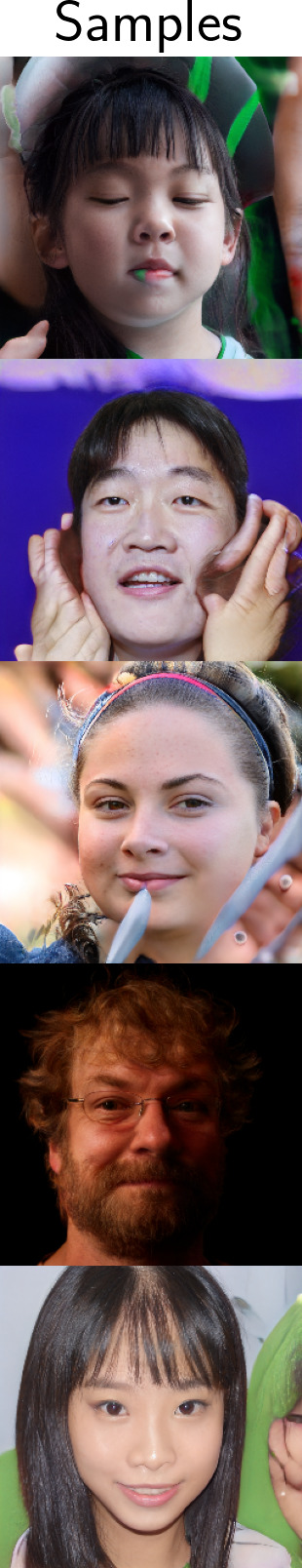}
		\caption{\\ Samples}
	\end{subfigure}%
	\begin{subfigure}{\samplecomparisonwidthfid}
		\centering
		\includegraphics[height=\samplecomparisonheight]{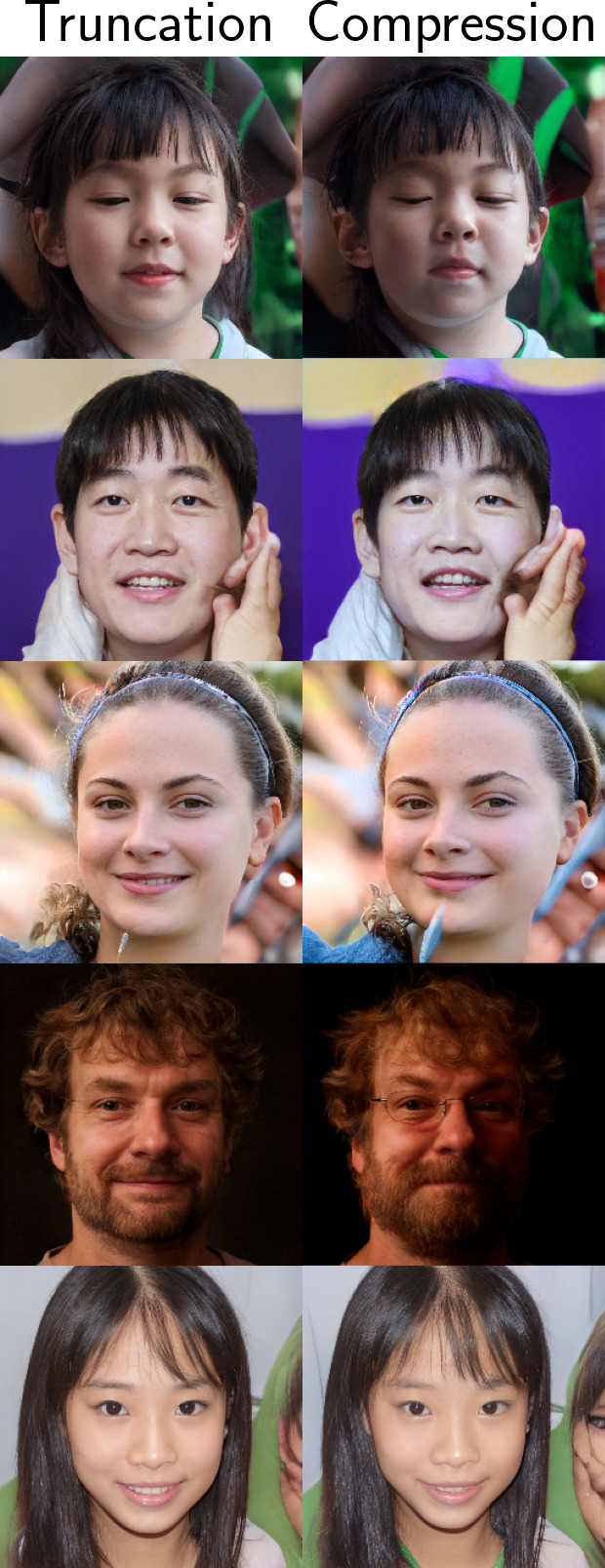}
		\caption{\\FID = 4}
	\end{subfigure}%
	\begin{subfigure}{\samplecomparisonwidthfid}
		\centering
		\includegraphics[height=\samplecomparisonheight]{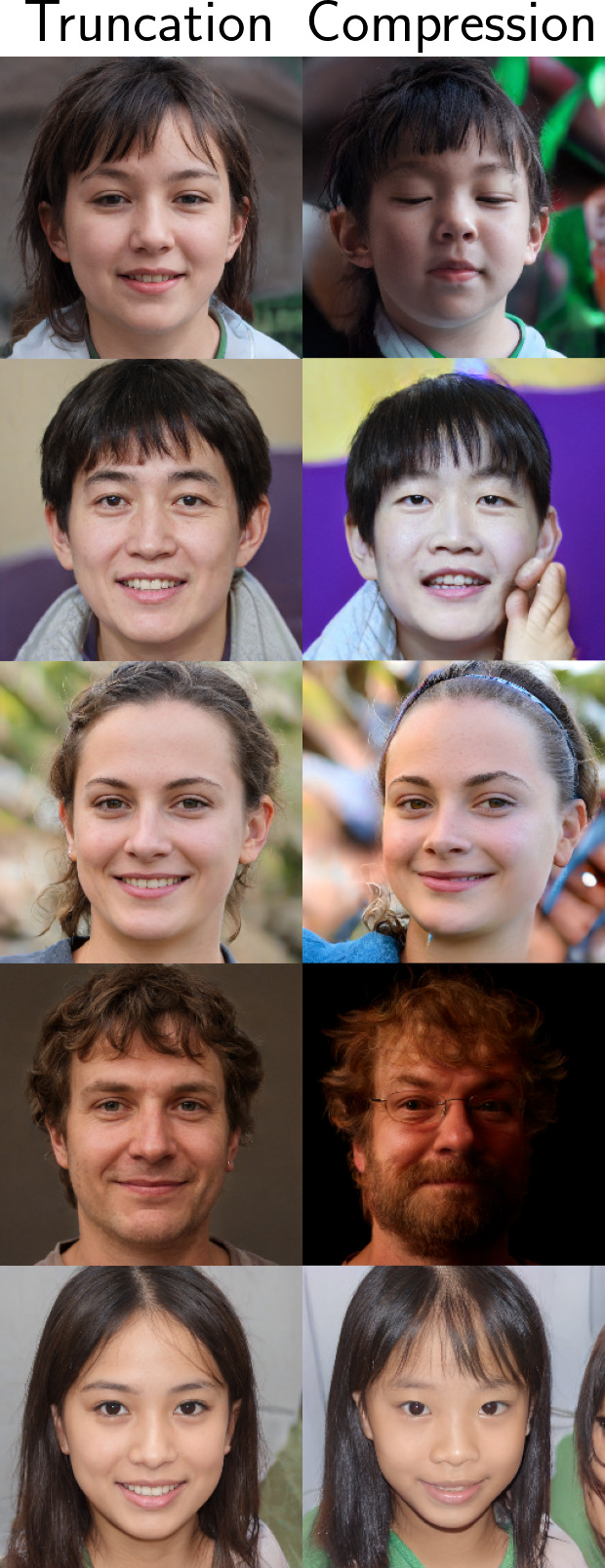}
		\caption{\\FID = 8}
	\end{subfigure}
	\caption{Comparison of correction methods. (a) The impact of the truncation parameter $\psi$ and the compression parameter $\tau$ on the FID is very similar. (b) At the same FID, our method better preserves facial identity. (c) shows a random selection of samples, (d-e) compare visual quality at the same FID values for truncation and compression (left and right columns, respectively). Both methods are effective at removing artifacts, but our method better preserves diversity.}
	\label{fig:truncation_vs_compression}
\end{figure}

\subsection{Results}
To be able to compare our proposed compression with the commonly used truncation method, we first need to align their parameters ($\psi$ for truncation and $\tau$ for compression).
Here, we use the Fréchet Inception Distance~\cite{Heusel:2017:FID} (FID, computed using 50K samples) to the training set as an alignment target, that is, we align $\psi$ and $\tau$ so that the corresponding corrected outputs have comparable FIDs to the training set.
Figure~\ref{fig:truncation_vs_compression}(a) illustrates the concept, and shows that the overall shape of the curve is similar.
Figure~\ref{fig:truncation_vs_compression}(c-e) show visual example for truncation and our compression method at matched FIDs. As we move away from the training distribution, truncation causes diversity to decrease. In the case of compression, on the other hand, the distance to the training set is caused by a lack of artifacts.
Interestingly, we note that for the same FID, our images are subjectively closer to the uncorrected image.
This indicates that FID in itself is not a particularly good metric to evaluate the presence or absence of artifacts in the image generation process and is unable to distinguish for example real makeup from failed attempts to draw makeup.

\textbf{Measuring face distortion.}
To provide a different perspective on the effects of correction, we investigate how either truncation or compression changes the identity of generated faces.
We sample 1024 images of faces without correction and correct them using both methods, with parameters matched to the same FIDs.
We then project all images into a face embedding space using the \texttt{SE-ResNet-50-128D} model from VGGFace2~\cite{Cao:2018:VggFace2}, and compute the cosine similarity between the uncorrected and corrected images in the embedding space.
Figure~\ref{fig:truncation_vs_compression}(b) shows the results.
In case of truncation (red), image correction significantly changes the identity of the samples, and the similarity decreases.
In contrast, our method (blue) preserves identity better, even with strong corrections.
\section{Conclusion}
In this work, we propose to analytically describe the distribution of data in a learned latent space of a Generative Adversarial Network. We show how, under a simple nonlinear transformation (an element-wise leaky ReLU), the distribution in the latent space can be modeled as a Gaussian in closed form.
We have demonstrated the benefits in two scenarios.
First, we have shown that the Gaussian distribution can serve as an effective prior for the task of image inversion, which significantly increases the accuracy of the recovered latent vector.
Second, we have shown that our Gaussian model allows us to analyze the cause of artifacts in the image generation pipeline and improve upon the common ``truncation'' trick by compressing principal components with large magnitudes.
This effectively removes artifacts, while keeping a higher degree of visual diversity compared to truncation.

\section{Acknowledgements}
This work was supported by a grant from Intel Corp. and a GPU gift from NVIDIA.

\bibliographystyle{splncs04}
\bibliography{egbib}

\section{Appendix}
\subsection{Broader impact}
In this work, we investigate the computational properties of Generative Adversarial Networks, and their impact on inversion and output quality.
While our primary goal is to understand GANs from a computational standpoint, both applications carry undeniable risks.
First, improving the quality of inversion could be desirable from an artistic standpoint, but can also be problematic if the subject whose image is inverted does not give consent.
Since our method improves inversion quality, modified images will also be more realistic, and therefore potentially harder to distinguish from real images. Further work on detecting synthetic images is therefore paramount.
Similarly, artifacts are often indicative of synthetic images, and removing them might make such images harder to identify.

However, a number of positive impacts of the the methods presented here are also possible.
For example, explicitly characterizing the distribution on the latent space could make it possible to develop better methods to detect synthetic images, by evaluating whether an inversion is likely to come from the latent distribution.
Second, increasing inversion quality could lead to a shift in focus in research on GANs, concentrating more on aspects of representation learning and less on generating better and more accurate fake images.
Lastly, we present a method to reduce artifacts on synthetic images without sacrificing diversity. If a GAN is used to generated synthetic training data for a downstream application, a lack in diversity is problematic, and can lead for example to systems that achieve much better classification performance for subjects with lighter skin color than for subjects with darker skin color.
Using our method as an alternative to the truncation trick could be a way to circumvent this issue.
\subsection{Hyperparameter tests}
\label{sec:hyperparameter}
Here, we give results for interpolation performance (similar to Figure 2 in the primary paper) for different prior weights $\lambda$.
While the best parameter differs between datasets, for consistency we choose $\lambda = 10^{-4}$. As can be seen from the graphs, this value provides a good compromise between interpolation performance and reconstruction performance at the endpoints (\ie raw inversion performance).
In Fig.~\ref{fig:appendix_errors_wplus}(a), for example, it can be seen that $\lambda = 10^{-3}$ (red curve) often provides better performance at the interpolation midpoint; however, in particular for Cars and Horses, this choice would provide a significantly worse inversion quality. 

\newpage
\newcommand{\ablationgraphwidth}{0.48\textwidth}
\begin{figure}
	\centering
	\captionsetup[subfigure]{justification=centering}
	\begin{subfigure}[t]{\ablationgraphwidth}
	\centering
		\includegraphics[width=\textwidth]{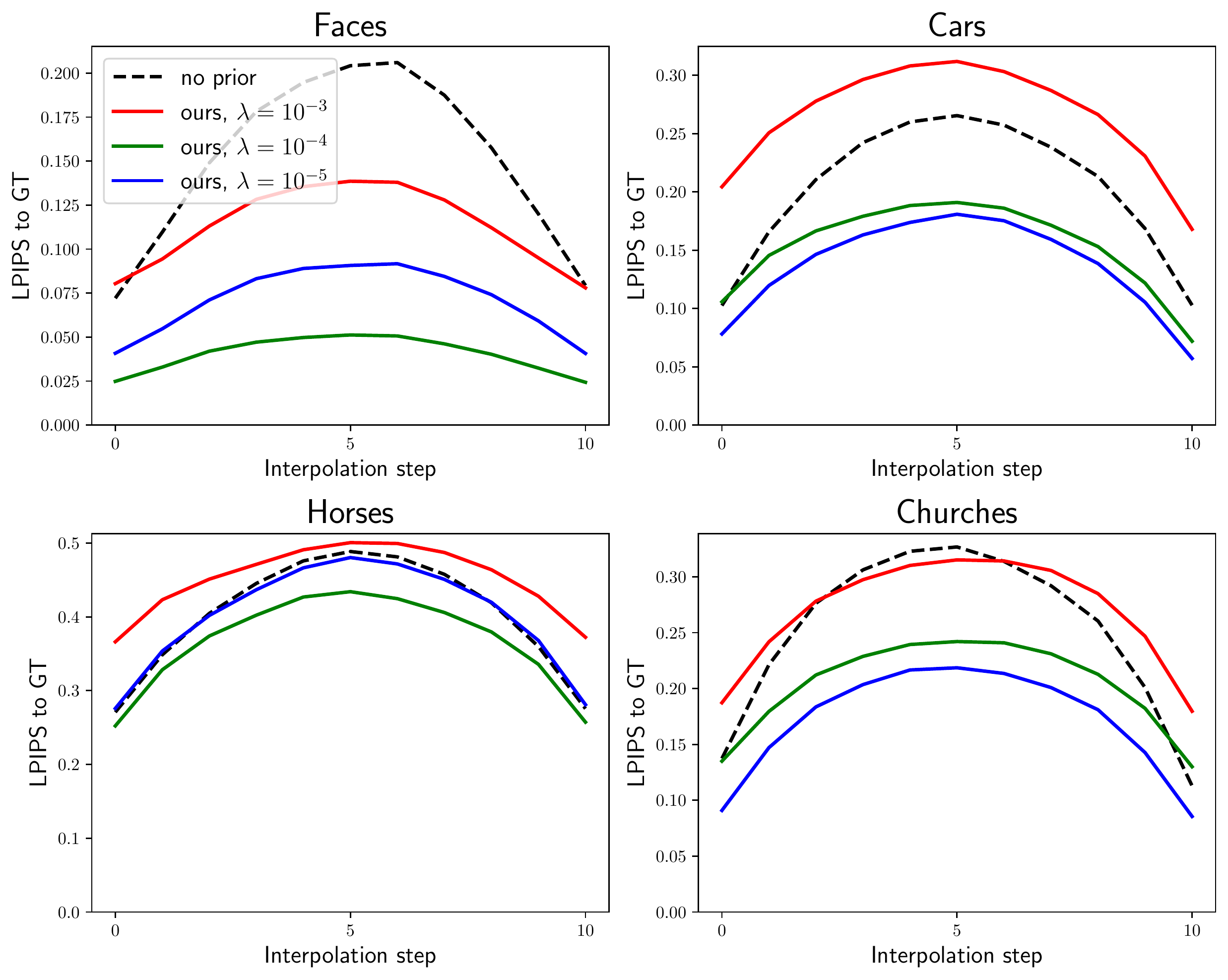}
		\caption{Reconstruction errors for images \\ when optimizing to $\mathcal{W}$.}
	\end{subfigure}
	\hspace{0.1in}
	\begin{subfigure}[t]{\ablationgraphwidth}
	\centering
		\includegraphics[width=\textwidth]{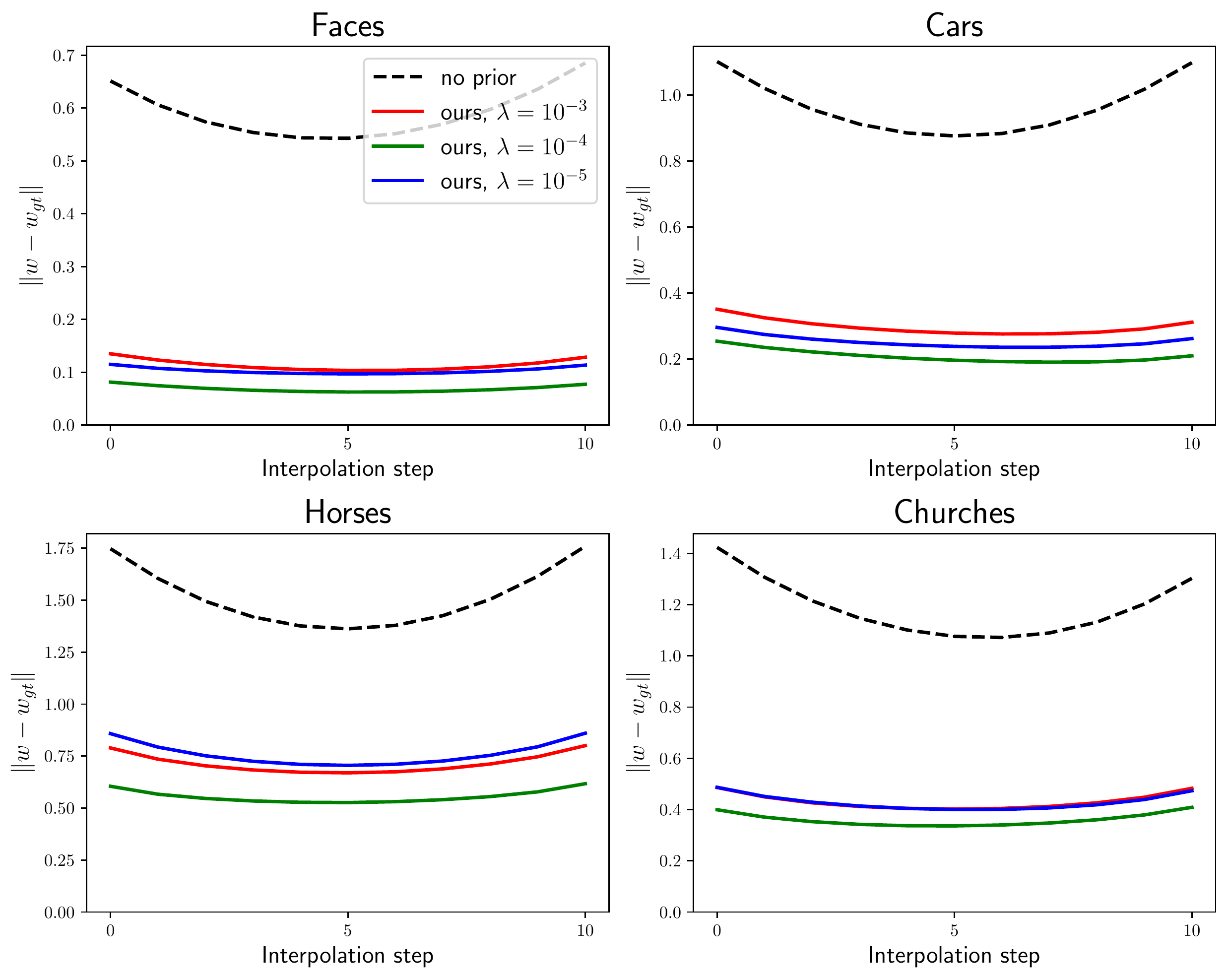}
		\caption{Reconstruction errors for latents \\ when optimizing to $\mathcal{W}$.}
	\end{subfigure} \vspace{0.15in}
	\caption{Average image and latent reconstruction errors when optimizing in $\mathcal{W}$ for different values for prior weight $\lambda$. The dashed line indicates the results when using an unconstrained optimization as described in~\cite{Karras_2019_Styleganv2}.}
	\label{fig:appendix_errors_w}
\end{figure}

\begin{figure}
	\centering
	\captionsetup[subfigure]{justification=centering}
	\begin{subfigure}[t]{\ablationgraphwidth}
	\centering
		\includegraphics[width=\textwidth]{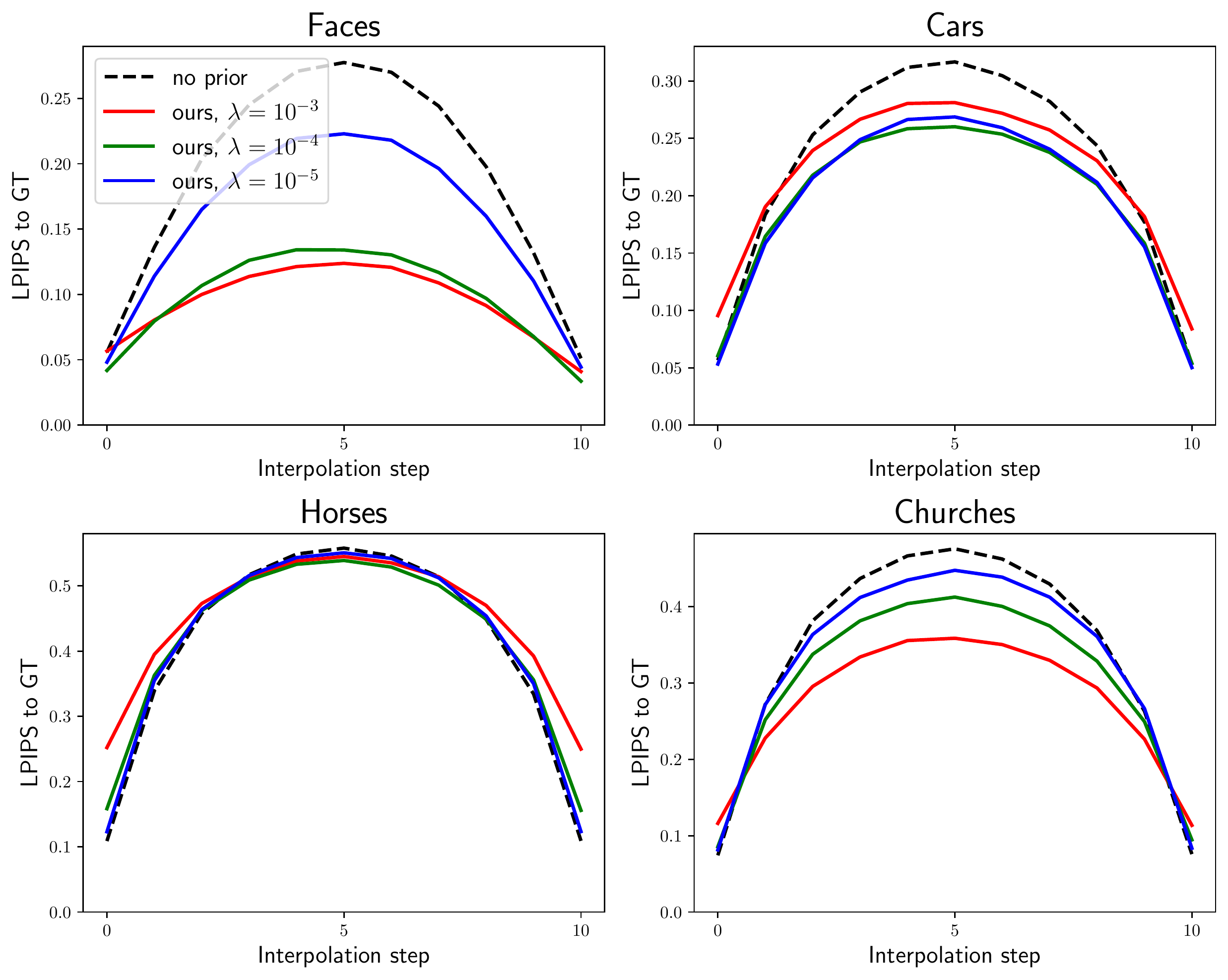}
		\caption{Reconstruction errors for images \\ when optimizing to $\mathcal{W}^{+}$.}
	\end{subfigure}
	\begin{subfigure}[t]{\ablationgraphwidth}
	\centering
		\includegraphics[width=\textwidth]{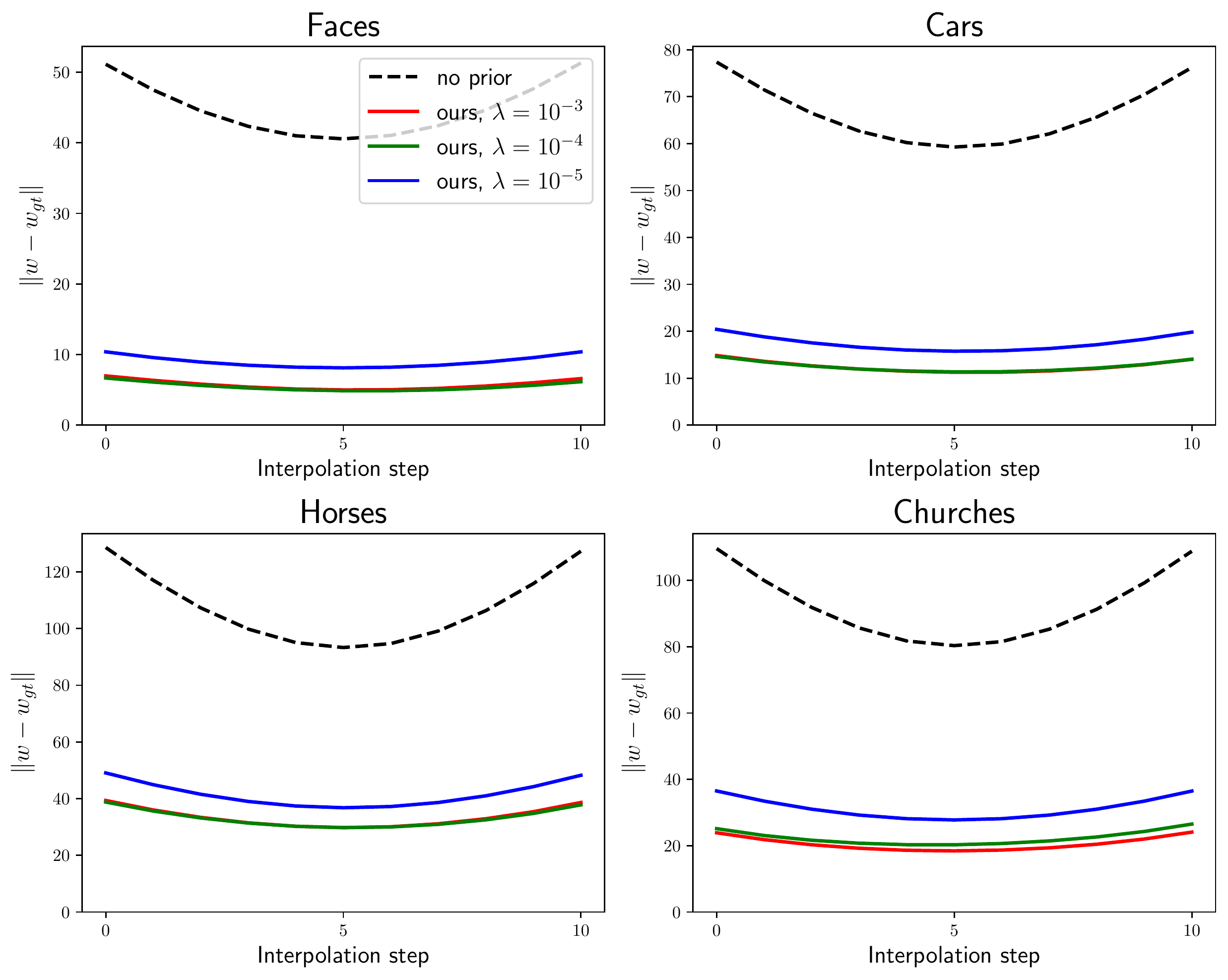}
		\caption{Reconstruction errors for latents \\ when optimizing to $\mathcal{W}^{+}$.}
	\end{subfigure} \vspace{0.15in}
	\caption{Average image and latent reconstruction errors when optimizing in $\mathcal{W}^{+}$ for different values for prior weight $\lambda$. The dashed line indicates the results when using an unconstrained optimization as described in~\cite{Karras_2019_Styleganv2}.}
	\label{fig:appendix_errors_wplus}
\end{figure}

\clearpage
\subsection{Additional interpolation results}
\label{sec:interpolationexamples}

\newcommand{\interpolwidth}{0.98\textwidth}
\begin{figure}[h]
	\centering
	\includegraphics[width=\interpolwidth]{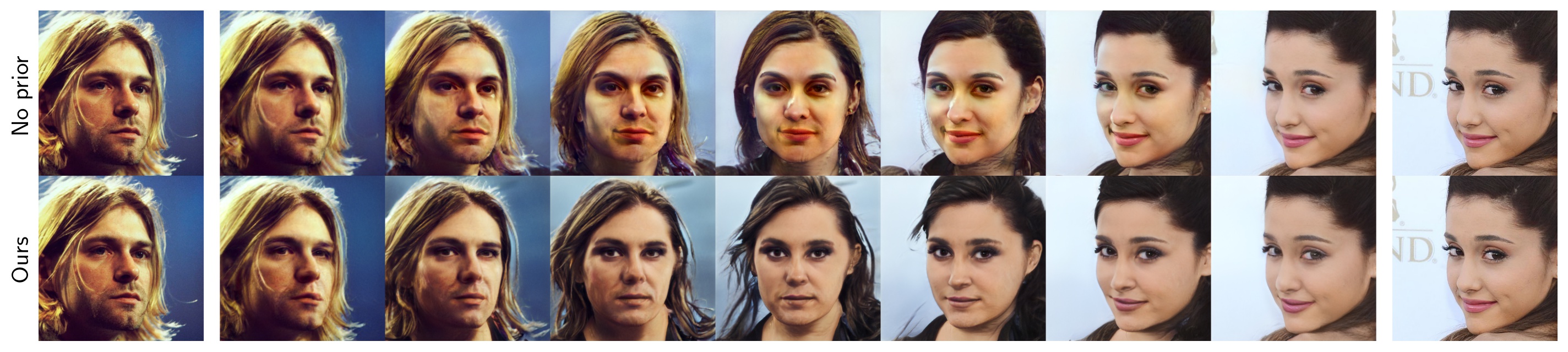}
		\includegraphics[width=\examplewidth]{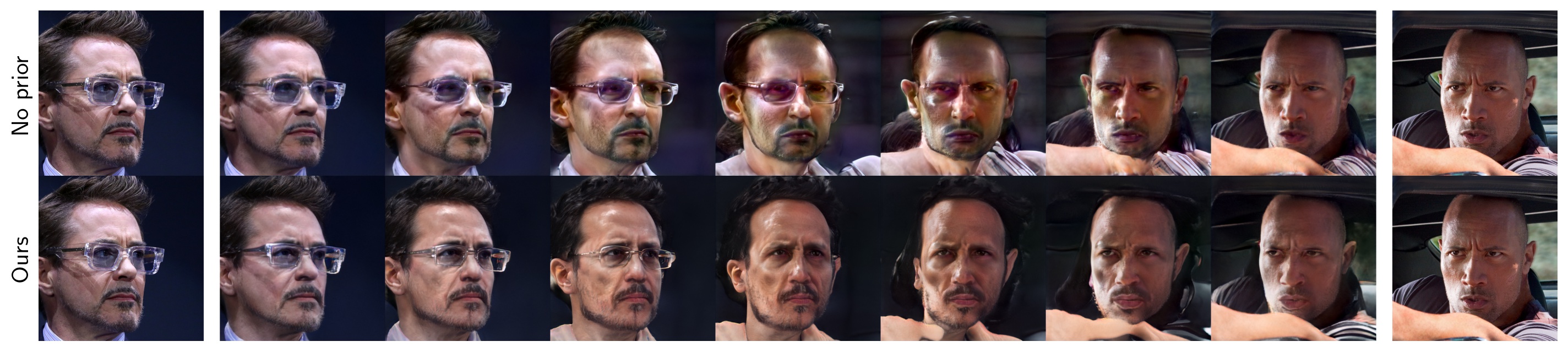}	
	\includegraphics[width=\interpolwidth]{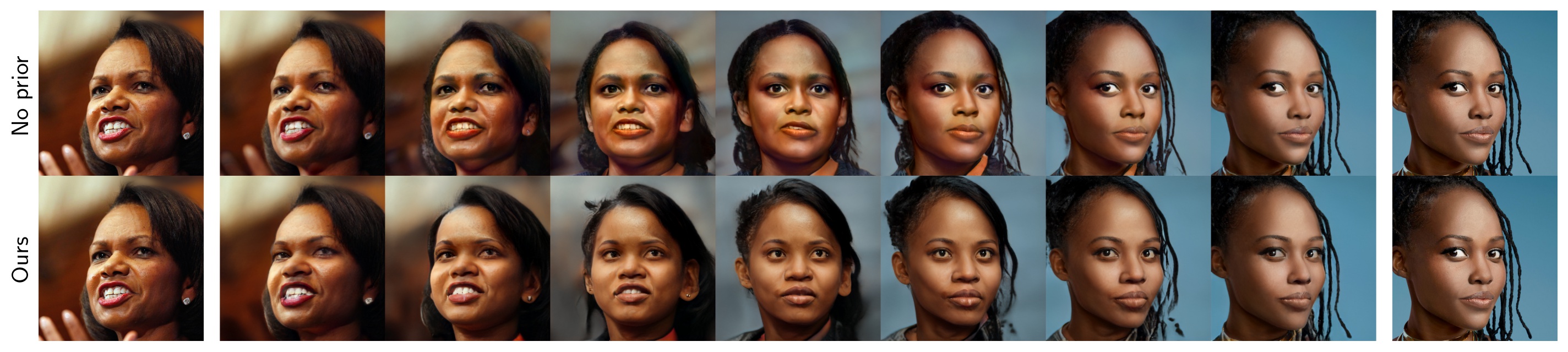}
	\includegraphics[width=\interpolwidth]{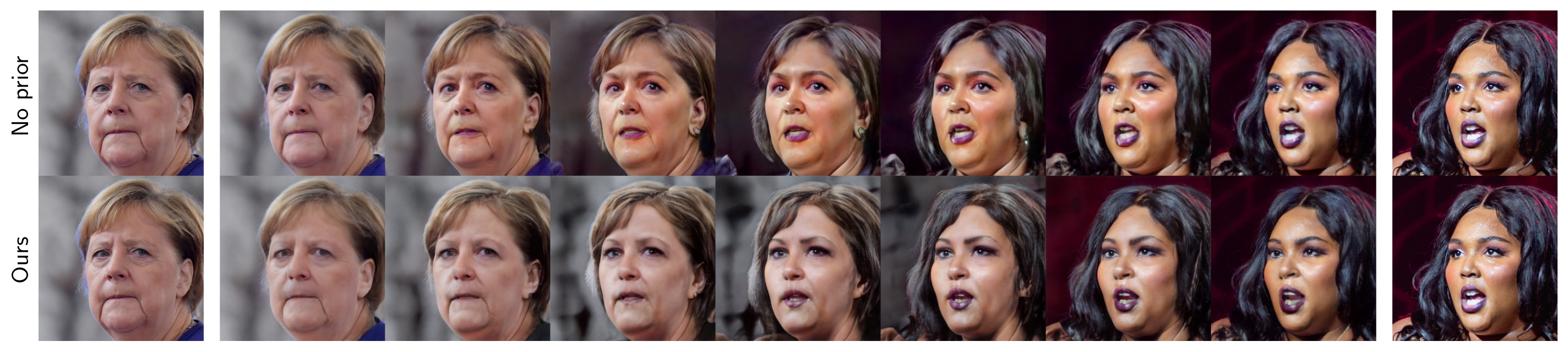}
	\includegraphics[width=\interpolwidth]{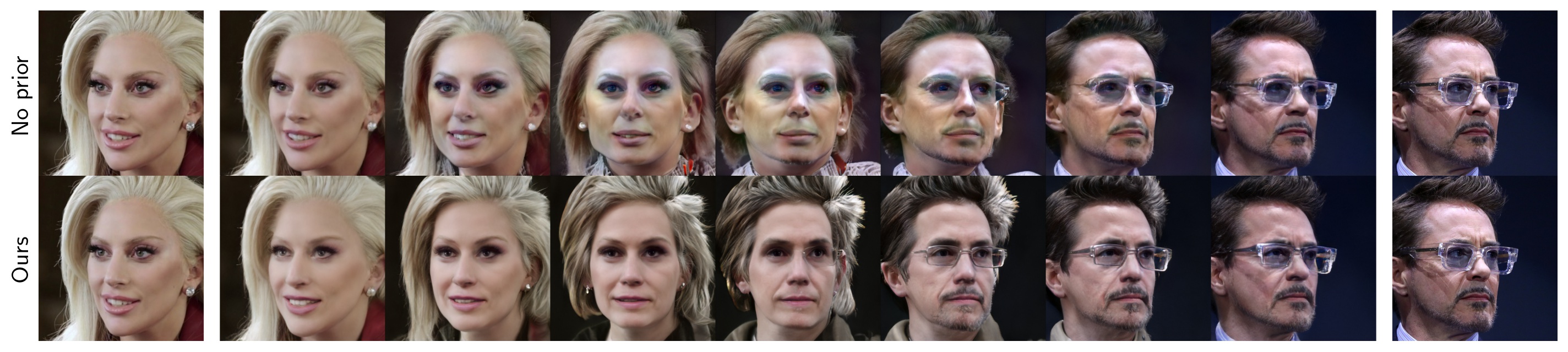}
	\includegraphics[width=\interpolwidth]{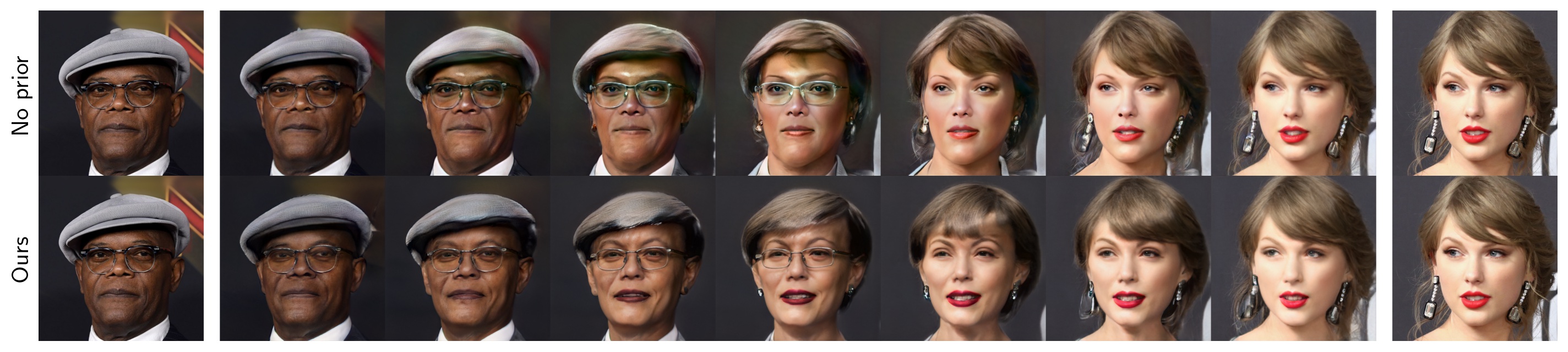}
	\caption{More examples for face interpolations. In each example, the top row shows an unconstrained optimization as described in~\cite{Karras_2019_Styleganv2}, but to $\mathcal{W}^{+}$~\cite{Abdal_2019_ICCV_Image2Stylegan}. The bottom row shows the results when using our proposed prior on the Gaussianized latent space $\mathcal{V}^{+}$.}
	\label{fig:interpolations_face}
\end{figure}

\renewcommand{\interpolwidth}{0.99\textwidth}
\begin{figure}[h]
	\centering
	\includegraphics[width=\interpolwidth]{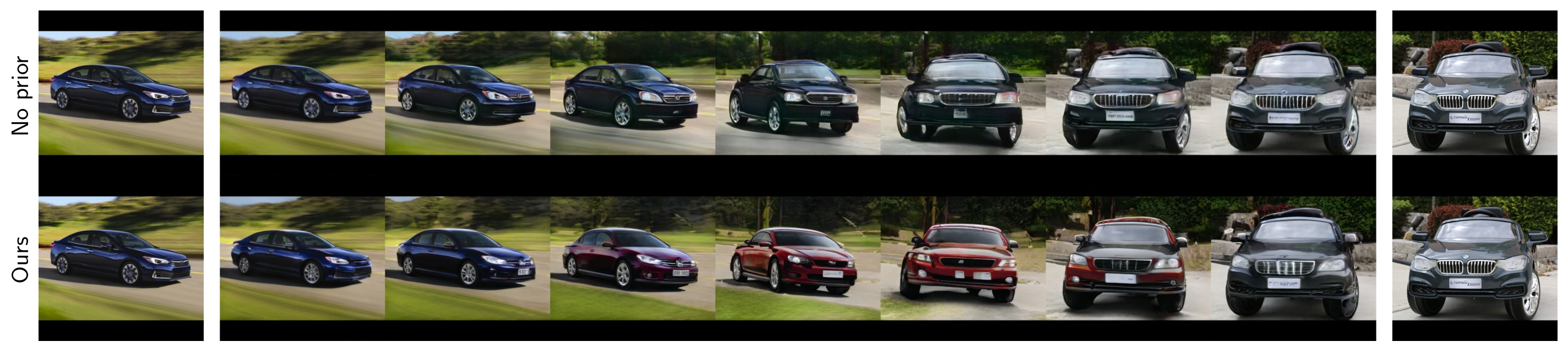}
	\includegraphics[width=\interpolwidth]{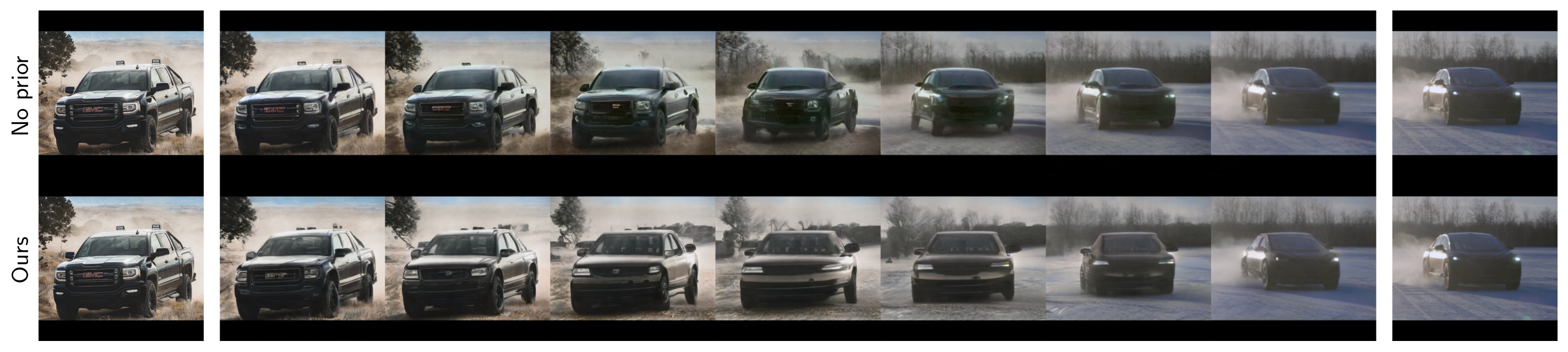}
	\includegraphics[width=\interpolwidth]{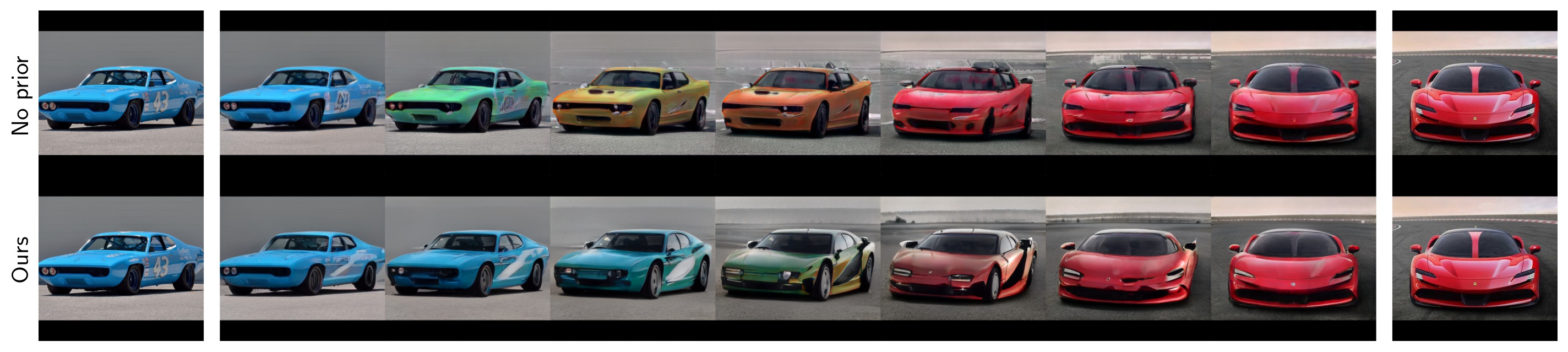}
	\includegraphics[width=\interpolwidth]{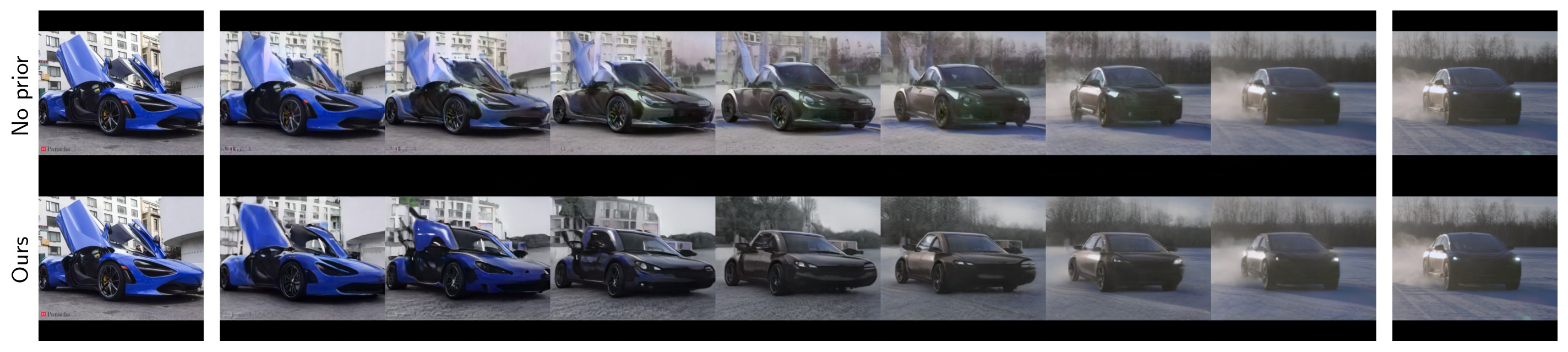}
	\includegraphics[width=\interpolwidth]{figures/03_interpolations/interpolation_cars_car18_car15.jpg}
	\caption{Examples for car interpolations. In each example, the top row shows an unconstrained optimization as described in~\cite{Karras_2019_Styleganv2}, but to $\mathcal{W}^{+}$~\cite{Abdal_2019_ICCV_Image2Stylegan}. The bottom row shows the results when using our the proposed prior on the Gaussianized latent space $\mathcal{V}^{+}$.}
	\label{fig:interpolations_car}
\end{figure}

\begin{figure}[h]
	\centering
	\includegraphics[width=\interpolwidth]{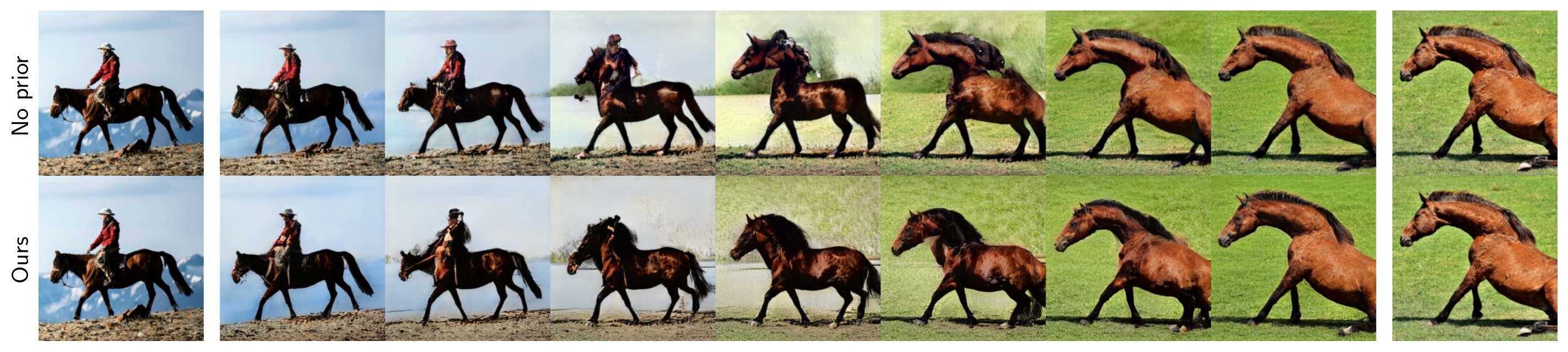}
	\includegraphics[width=\interpolwidth]{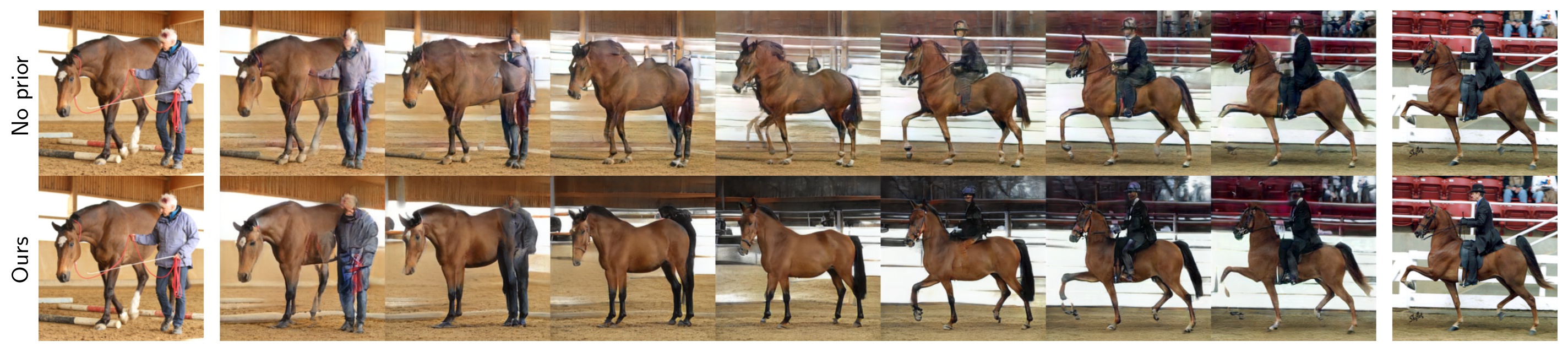}
	\includegraphics[width=\interpolwidth]{figures/03_interpolations/interpolation_horses_horse06_horse18.jpg}
	\includegraphics[width=\interpolwidth]{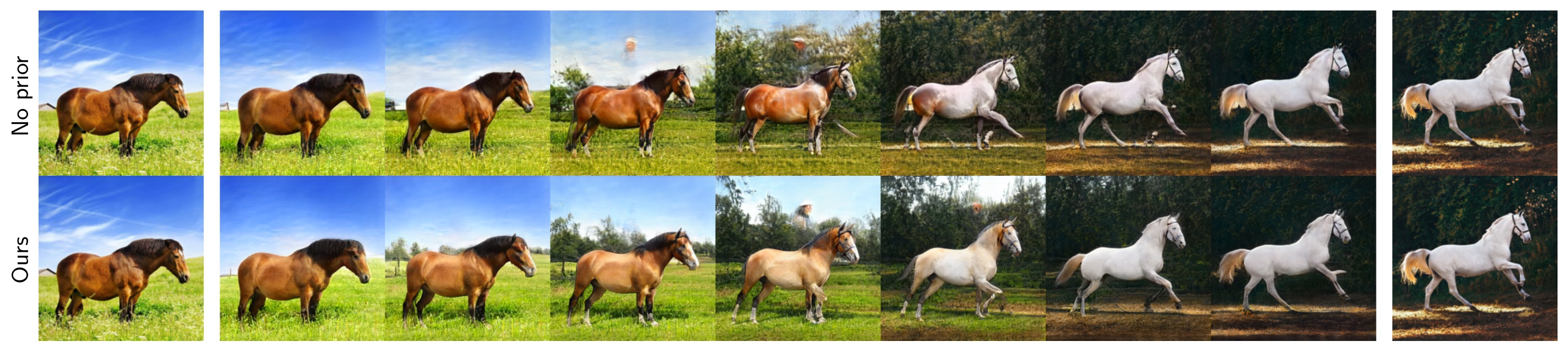}
	\includegraphics[width=\interpolwidth]{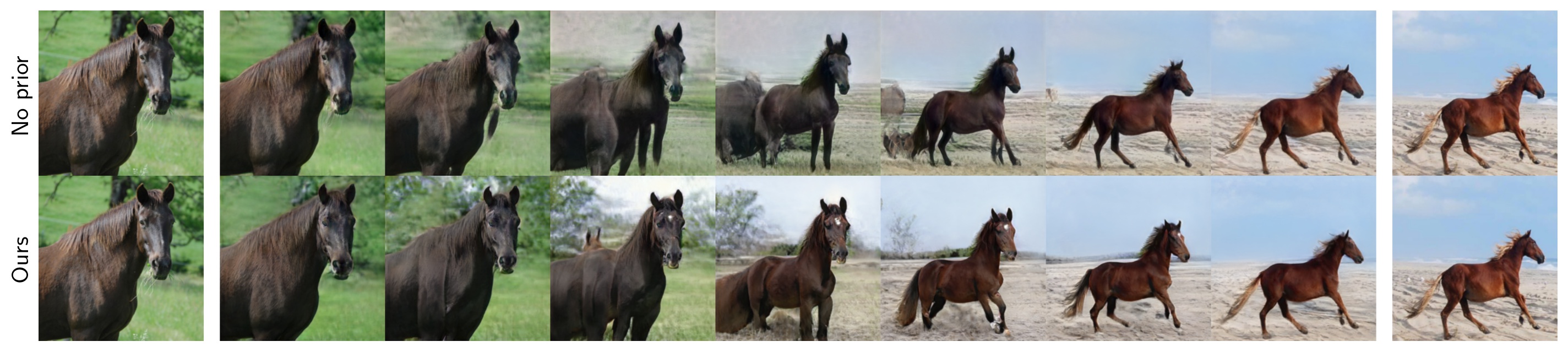}
	\caption{Examples for horse interpolations. In each example, the top row shows an unconstrained optimization as described in~\cite{Karras_2019_Styleganv2}, but to $\mathcal{W}^{+}$~\cite{Abdal_2019_ICCV_Image2Stylegan}. The bottom row shows the results when using our the proposed prior on the Gaussianized latent space $\mathcal{V}^{+}$.}
	\label{fig:interpolations_horse}
\end{figure}

\begin{figure}[h]
	\centering
	\includegraphics[width=\interpolwidth]{figures/03_interpolations/interpolation_churches_church02_church01.jpg}
	\includegraphics[width=\interpolwidth]{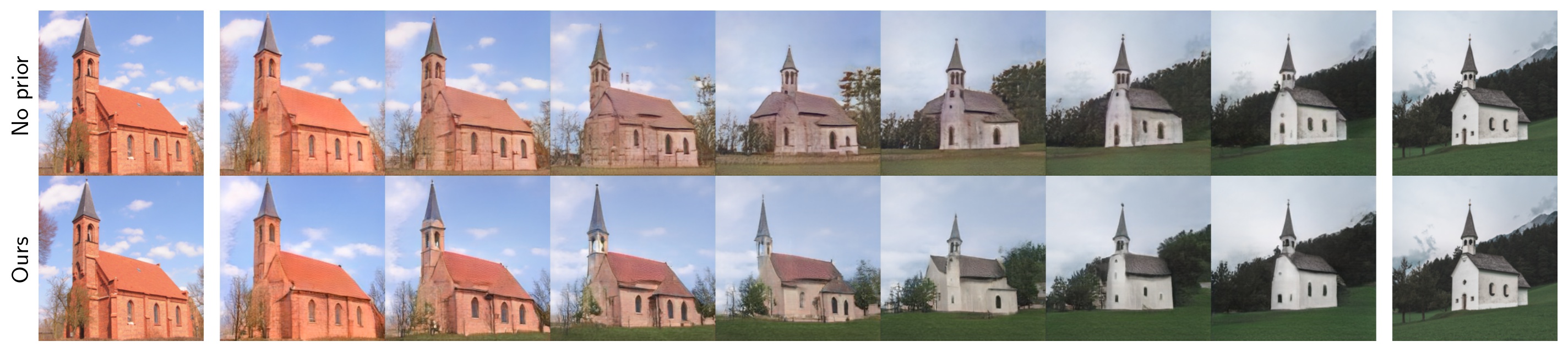}
	\includegraphics[width=\interpolwidth]{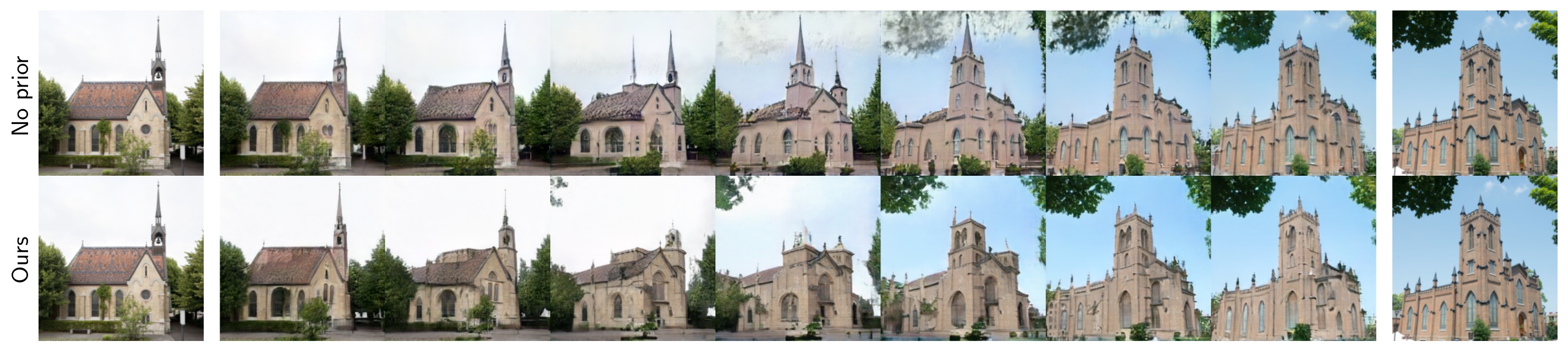}
	\includegraphics[width=\interpolwidth]{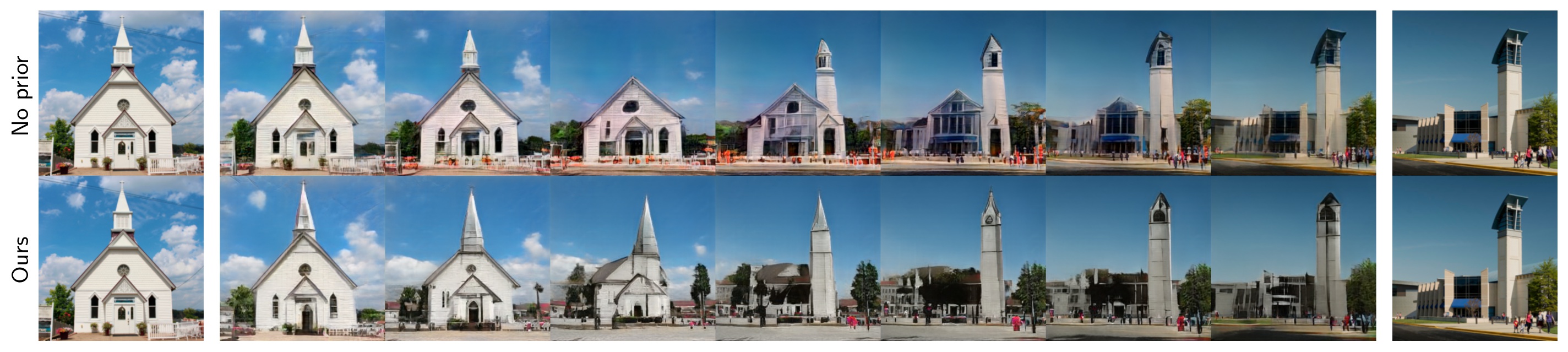}
	\includegraphics[width=\interpolwidth]{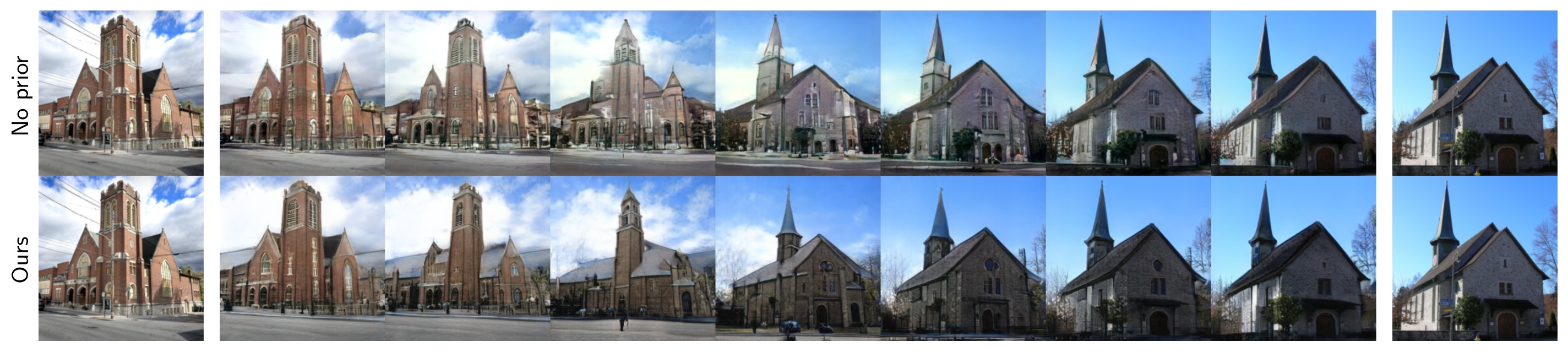}
	\caption{Examples for church interpolations. In each example, the top row shows an unconstrained optimization as described in~\cite{Karras_2019_Styleganv2}, but to $\mathcal{W}^{+}$~\cite{Abdal_2019_ICCV_Image2Stylegan}. The bottom row shows the results when using our the proposed prior on the Gaussianized latent space $\mathcal{V}^{+}$.}
	\label{fig:interpolations_church}
\end{figure}
\clearpage

\end{document}